\documentclass{article}

\PassOptionsToPackage{numbers,sort&compress}{natbib}
\usepackage[preprint]{neurips_2026}

\usepackage[utf8]{inputenc}
\usepackage[T1]{fontenc}
\usepackage{hyperref}
\usepackage{url}
\usepackage{booktabs}
\usepackage{multirow}
\usepackage{amsmath}
\usepackage{amsfonts}
\usepackage{amssymb}
\usepackage{nicefrac}
\usepackage{microtype}
\usepackage{graphicx}
\usepackage{wrapfig}
\usepackage{afterpage}
\usepackage{xcolor}

\newcommand{\finding}[1]{%
  \par\vspace{4pt}\noindent\fcolorbox{black!40}{black!4}{%
    \parbox{\dimexpr\linewidth-2\fboxsep-2\fboxrule\relax}{%
      \textbf{Finding.}\enspace #1%
    }%
  }\par\vspace{4pt}%
}

\title{What Do Evolutionary Coding Agents Evolve?}

\author{%
  \textbf{Nico Pelleriti$^{1,2}$ \quad Sree Harsha Nelaturu$^{1}$ \quad Zhanke Zhou$^{3}$ \quad Zongze Li$^{3}$} \\[0.2em]
  \textbf{Max Zimmer$^{1}$ \quad Bo Han$^{3,4}$ \quad Sebastian Pokutta$^{1,2}$} \\[0.4em]
  \normalfont $^{1}$Zuse Institute Berlin \quad $^{2}$Technical University of Berlin \\
  \normalfont $^{3}$Hong Kong Baptist University \quad $^{4}$RIKEN Center for Advanced Intelligence Project
}

\begin{document}

\maketitle

\begin{abstract}
Recent work pairs LLMs with evolutionary search to iteratively generate, modify, and select code using task-specific feedback. 
These systems have produced strong results in mathematical discovery and algorithm design, yet a fundamental question remains: what do they actually evolve? 
Progress is typically summarized by the best score a run reaches under a task-specific evaluator, but that score can reflect several different mechanisms: new algorithmic structure, re-tuning an existing strategy, recombining ideas already in the model's internal knowledge, or overfitting to the evaluator. 
Distinguishing these mechanisms requires inspecting the search process itself, not only its final outcome. 
We introduce EvoTrace, a dataset of evolutionary coding traces spanning four evolutionary frameworks, reasoning and non-reasoning models, and 16 tasks across mathematics and algorithm design. 
To analyze these traces, we develop EvoReplay, a replay-based methodology that reconstructs the local search states behind high-scoring solutions and tests controlled interventions, including adjusting constants, removing program components and substituting models or prompting contexts. 
We annotate every code edit in EvoTrace with one of nine recurring edit types using an LLM-as-judge pipeline validated against blind human re-annotation.
Across EvoTrace, most score gains come from a small subset of these edit types. We further find a deterministic cycling pattern: about 30\% of code lines added during search are byte-identical re-introductions of previously-deleted lines, present throughout nearly every run.
These results show that benchmark gains in evolutionary coding agents can arise from qualitatively different mechanisms, only some of which correspond to new algorithmic structure. EvoTrace enables more diagnostic evaluation of evolutionary coding agents beyond final benchmark scores.
\end{abstract}
\section{Introduction}
\label{sec:intro}

Large Language Model (LLM)-driven evolutionary code search has rapidly emerged as a promising paradigm for automated scientific and engineering discovery. In this setting, LLMs propose program mutations within search loops that are guided by executable feedback \citep{romera-paredes2024,novikov2025c,georgiev2025c,lange2025b,assumpcao2026,cemri2026,liu2026}.
This paradigm has produced strong results across mathematical construction, systems optimization, algorithm design, and GPU kernel engineering, including improved bounds for combinatorial problems, better packing constructions, compiler and scheduling heuristics \citep{romera-paredes2024,novikov2025c,georgiev2025c,cheng2025b,guo2025,cao2026,wiedemann2026}.
In this study, we define an \textbf{evolutionary coding agent} as a system with a task specification and executable evaluator, a population or archive of candidate programs, one or more LLMs that generate code mutations, recombinations, or refinements, and a search procedure that selects which programs and contexts feed future generations. A \textbf{search trace} is the full record produced by such a system: generated programs, scores, execution feedback, parent-child relations, prompts, model choices, and intermediate artifacts.

\begin{figure}[t]
    \centering
    \includegraphics[width=\linewidth]{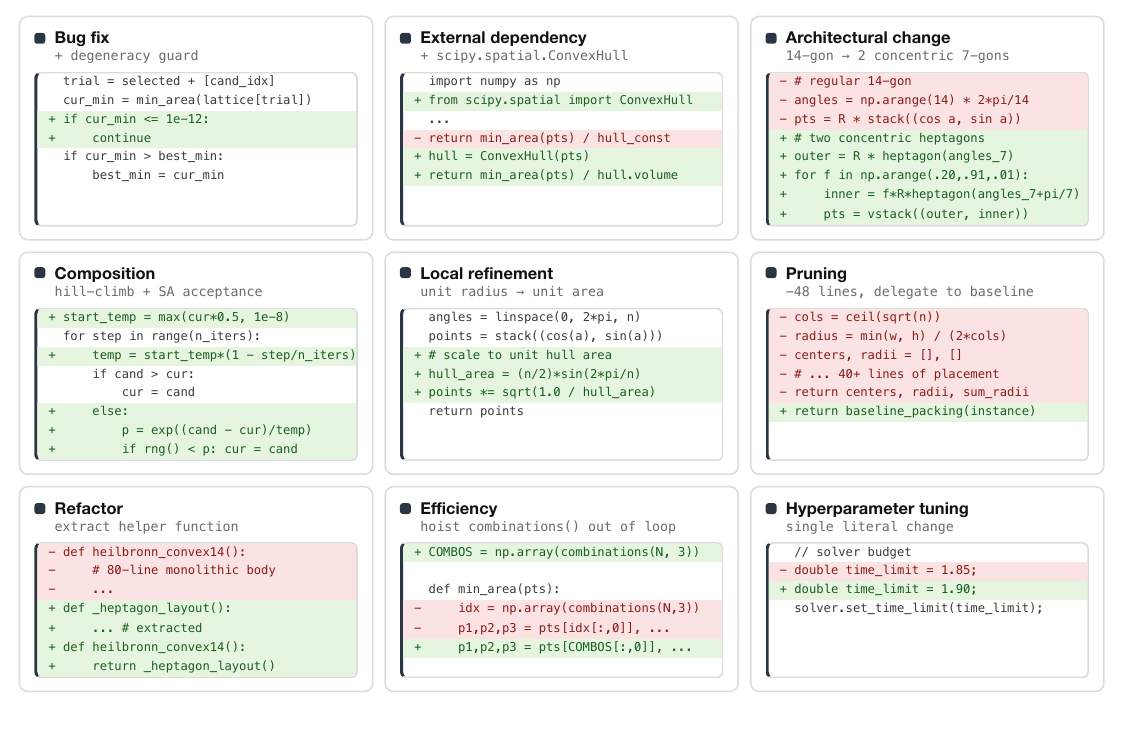}
    \caption{\textbf{A taxonomy of edits performed by evolutionary coding agents.} Each panel shows a representative parent--child diff (added lines in green, deleted lines in red) drawn from EvoTrace runs and labeled with one of nine recurring categories: \emph{Bug fix}, \emph{External dependency}, \emph{Architectural change}, \emph{Composition}, \emph{Local refinement}, \emph{Pruning}, \emph{Refactor}, \emph{Efficiency}, and \emph{Hyperparameter tuning}. The categories range from minimal numeric edits (a single literal change) to structural rewrites (replacing a 14-gon with two concentric heptagons), and they form the basis of the LLM-as-judge edit annotation used throughout the paper. Edits are typically multi-label; we examine prevalence and per-edit utility in \S\ref{sec:static-analysis}.}
    \label{fig:evolutionary_edits_taxonomy}
\end{figure}

Despite rapid interest and empirical progress, the internal dynamics of evolutionary coding agents remain poorly understood. Existing work often reports final best scores, aggregate success rates, or a handful of illustrative trajectories, but such endpoints obscure the pathways and mechanisms by which improvements arise. While evolutionary coding agents sometimes demonstrate clear advantages over baselines such as independent sampling, greedy refinement, or beam-style search, these gains are highly sensitive to choices in task design, initialization, evaluator specification, or model configuration.
There is consequently no clear consensus on what these systems are actually evolving: whether progress comes from discovering new program structure, tuning parameters of already-known strategies, recombining concepts already present in the model, or preserving early biases in the population.
This motivates the central question we address: \textit{What do evolutionary coding agents evolve, and how do their search dynamics produce improvements?}

To address this question, we introduce a dataset of evolutionary coding traces collected across multiple frameworks, models, and task families, covering mathematical constructions and algorithmic programming tasks. Rather than treating each run only by its final score, we study the full trajectory of generated programs: which solutions are explored, which lineages produce major improvements, and how stable those improvements are. The resulting dataset is intended to make evolutionary coding agents analyzable as dynamic systems, not just benchmark submissions.

Analyzing these traces is non-trivial: a single run may generate hundreds of unique programs, and each candidate can differ from its ancestors through non-local structural changes, small numerical edits, prompt-driven rewrites, or evaluator-specific hacks. To make runs comparable, we develop a unified trace representation together with an annotation and measurement pipeline. The representation exposes the search graph, candidate programs, evaluations, and lineage information, enabling analyses of population structure, diversity, best-lineage stability, counterfactual model or context changes, and the decomposition of structural versus parametric gains.

We use this framework as a diagnostic tool for understanding both progress and failure in evolutionary coding.
Our analysis covers four diagnostics: static measures of program complexity and lineage utilization (\S\ref{sec:static-analysis}), deterministic detection of cycling, the re-introduction of previously-deleted lines (\S\ref{sec:cycling}), 
\begin{wrapfigure}{r}{0.65\textwidth}
    \centering
    \vspace{-0.8em}
    \includegraphics[width=0.65\textwidth]{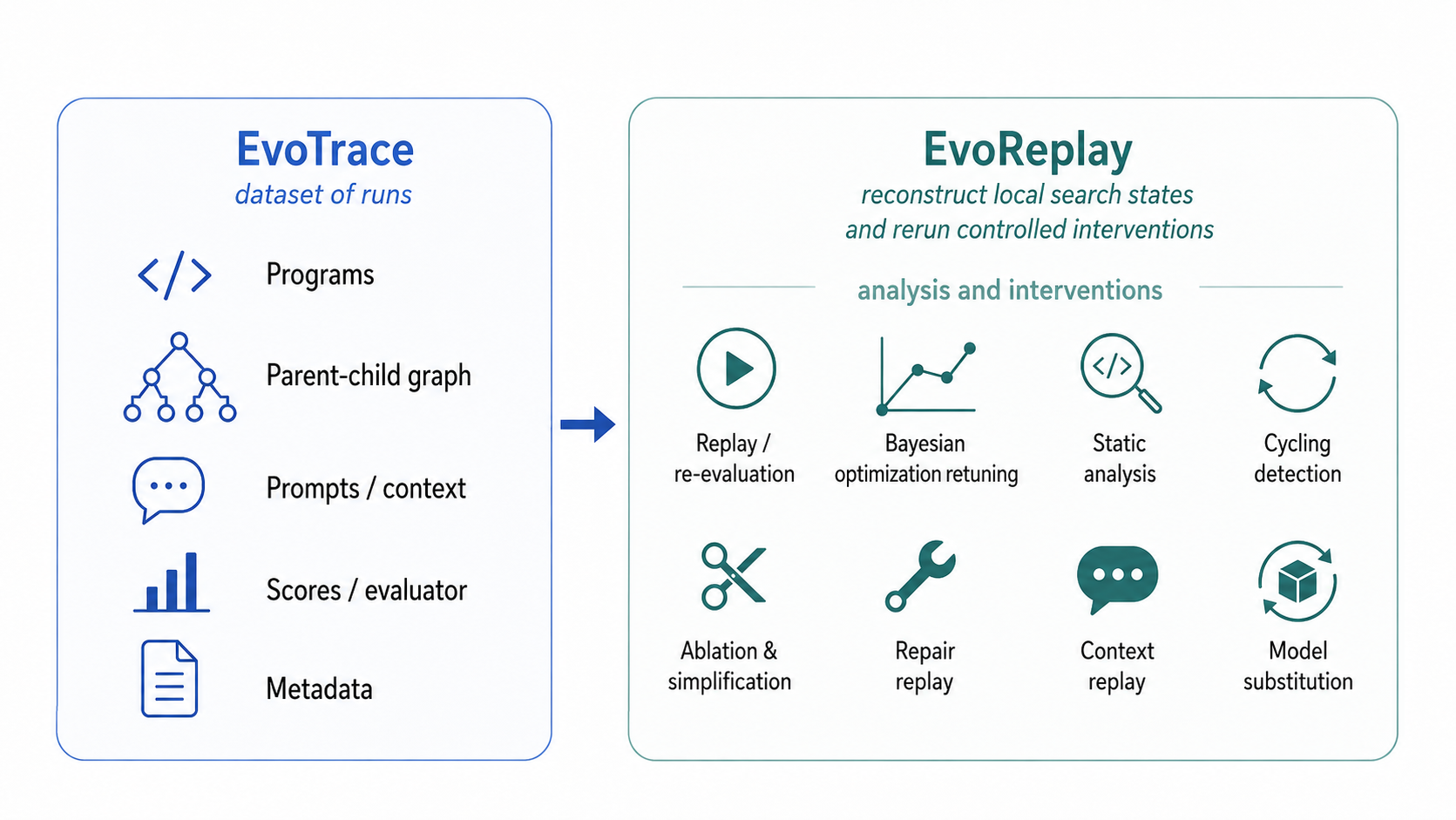}
    \vspace{-1.2em}
    \caption{\textbf{EvoTrace and EvoReplay.} EvoTrace records each evolutionary run as a structured object: programs, parent--child graph, prompts and context, scores, and evaluator metadata. EvoReplay reconstructs local search states from these traces and reruns controlled interventions, including same-prompt replay, Bayesian-optimization retuning, static analysis, cycling detection, ablation, repair, context substitution, and model substitution.}
    \label{fig:overview}
    \vspace{-0.8em}
\end{wrapfigure}replay-based stability tests on score-improving edits (\S\ref{sec:replay}), and a tuning-gap baseline that runs Bayesian optimization over the hyperparameters of a single program (\S\ref{sec:tuning-gap}).

These diagnostics are meant to be practical: the same trace representation and measurements can be integrated into existing open-source evolutionary coding frameworks to reveal whether a run is discovering new algorithmic structure, retuning known patterns, or becoming trapped by its own history. Overall, our results suggest that progress in evolutionary coding is not explained by final scores alone, but depends jointly on the task, the search procedure, the evaluator, and the underlying model family. By measuring the dynamics of search traces, we take a first step toward a systematic understanding of what these agents change over time, which changes matter, and why some runs keep improving while others stall.

\paragraph{Contributions.}
We contribute two artifacts and a set of trace-level findings that use them.
\textbf{(1)~EvoTrace}\footnote{Available at \url{https://huggingface.co/datasets/ZIB-IOL/EvoTrace}.}, a dataset of $121$ evolutionary coding-search runs across four frameworks on $16$ benchmarks spanning Python mathematical constructions and C++ competitive programming problems, with $10{,}672$ unique programs, $18{,}400$ LLM calls, full parent--child graphs, prompts and contexts, scores, and evaluator metadata, normalized into a unified replayable schema (\S\ref{sec:dataset}).
\textbf{(2)~EvoReplay}\footnote{Available at \url{https://github.com/ZIB-IOL/EvoReplay}.}, a methodology and accompanying open-source package that reconstructs local search states from EvoTrace traces and reruns controlled interventions (same-prompt replay, Bayesian-optimization retuning, static analysis, deterministic cycling detection, ablation, repair, and context or model substitution), so that mechanistic claims about a search trajectory can be tested rather than asserted (\S\ref{sec:evoreplay}).
\textbf{(3)~Findings} from applying these tools to characterize how evolutionary coding agents actually behave: which edit types drive score gains, how often runs end up adding back code they had previously deleted (a cycling pattern present throughout the trajectory in nearly all runs), how reliably score-improving programs can be reproduced by re-running the same prompt, how well public scores generalize to held-out evaluation on competitive programming tasks, and how much of a math run's headline gain is recoverable by tuning the hyperparameters of a single mid-run program (\S\ref{sec:experiments}).

\section{Related Work}

\subsection{LLM-Guided Evolutionary Coding Approaches}
LLMs can act as mutation operators inside evolutionary loops over executable programs.
FunSearch \citep{romera-paredes2024} and AlphaEvolve \citep{novikov2025c,georgiev2025c} established this paradigm, and a growing collection of open-source frameworks now extend it, including OpenEvolve \citep{openevolve}, GEPA \citep{agrawal2026}, ShinkaEvolve \citep{lange2025b}, GigaEvo \citep{khrulkov2025}, CodeEvolve \citep{assumpcao2026}, the FM Agent \citep{li2026}, and AIDE \citep{jiang2025a}.
A second wave targets the search procedure itself by adapting strategies, models, or signals during the run \citep{liu2026,cemri2026,yan2026,ray2026,chen2026b}, or by changing the unit of evolution to solution spaces, strategies, skills, prompt groups, populations, or concept trees \citep{zhai2025,luo2026,ye2026,li2025c,zhang2025d,leleu2026,pourcel2026,wang2025a,yuksekgonul2026,ma2025,singhal2026,liu2026selfplayevolvesselfsyntheticpipeline}.

A particularly active line of work targets GPU kernel optimization, where wall-clock runtime provides a tight reward signal \citep{guo2025,wiedemann2026,cao2026,du2026,nichols2026,wei2025a,su2026,cheng2025a,cheng2025b}.
These frameworks have also been applied to compiler heuristics, computer architecture, cosmology, swarm-intelligence design, symbolic regression, retrieval, recommendation, hyperparameter optimization, code optimization, agentic reasoning, autonomous data science, and broader scientific discovery \citep{chen2026a,gupta2026,li2026a,cen2025,song2025,nian2026,wang2026a,ferreira2026,hu2026,zhou2026b,yang2025a,weng2026,skydiscover2026}.
This line builds on classical evolutionary computation and quality-diversity search \citep{koza1994,jaderberg2017,qian2024,nadizar2025,hughes2024}; because the evolved objects are programs, recent work develops program-aware notions of diversity and similarity \citep{friedman2023,zhang2026c}.
Our work is complementary: rather than proposing another framework, we analyze traces from four of them (OpenEvolve, GEPA, ShinkaEvolve, EvoX) to study the mechanisms by which programs improve, stagnate, diversify, or collapse.

\subsection{Analyzing Evolutionary Coding Agents}
A separate body of work studies evolutionary coding systems themselves.
Surveys situate the broader space \citep{fang2025,xie2025b}, benchmarks evaluate agents across ML competitions, long-horizon algorithm engineering, frontier research science, and production deployments \citep{chan2025b,imajuku2025,lupidi2026,ye2026a,pan2026,zheng2026a}, and \citet{gideoni2026} show that simple baselines can match elaborate evolutionary pipelines.
Closer to our methodology, several papers analyze search behavior rather than only final scores: trajectory analyses \citep{zhang2026d}, fitness-landscape characterization \citep{liu2025a}, failure modes of iterative LLM optimization \citep{nie2026}, exploration deficits \citep{pan2025}, output homogeneity \citep{jiang2025}, emergent risks in self-evolving agents \citep{shao2026}, and a taxonomy of multi-agent failures \citep{cemri2025}.
Our work adopts a similar diagnostic perspective but focuses on full evolutionary coding traces, which we use to characterize how populations evolve, how improvement propagates through lineages, and how search dynamics produce both successes and failure modes.

\section{EvoTrace}
\label{sec:dataset}

To understand what evolutionary coding agents evolve, we construct \emph{EvoTrace}, a dataset of structured search traces from LLM-driven evolutionary coding systems.
EvoTrace contains the artifacts needed to inspect and replay parts of the search process: generated programs, parent-child relations, prompts and retrieved context, evaluator outputs, scores, execution logs, and environment metadata.
The dataset covers mathematical constructions and competitive programming tasks.
These domains were chosen to capture distinct forms of code improvement: mathematical tasks reward new search algorithms, while competitive programming tasks require generated programs to compile and pass external judging, with limited access to the evaluator.

Different frameworks also log different parts of the search state, making direct cross-system comparison difficult.
To address these issues, EvoTrace treats evolutionary search traces not merely as logs to annotate after the fact, but as structured computational objects that can be normalized, replayed, and intervened on. The collection and replay infrastructure is built on top of SkyDiscover \citep{skydiscover2026}, a flexible framework for AI-driven scientific and algorithmic discovery; we extend it with a unified cross-backend schema, replay environments, and the analysis tooling described in \S\ref{sec:evoreplay}.

\subsection{Data Collection Across Tasks, Frameworks, and Models}
\label{sec:dataset_collection}

EvoTrace covers 16 different tasks across two language--domain pairs: 6 Python mathematical-discovery tasks (circle packing, Heilbronn placement, autocorrelation and uncertainty inequalities, signal processing) and 10 C++ competitive programming problems from ALE-bench Lite \citep{imajuku2025} (AtCoder Heuristic Contest), each with a judge-defined score. We measure four evolutionary coding systems (Table~\ref{tab:frameworks}): OpenEvolve \citep{openevolve}, GEPA \citep{agrawal2026}, EvoX \citep{liu2026}, and ShinkaEvolve \citep{lange2025b}.
They share the propose--evaluate--feedback pattern but differ in selection, context, diversity, and adaptation. 
We employ 5 different LLMs (Table~\ref{tab:models}) to generate the mutations, and 100 search iterations per run, with a total of 121 runs and over 10,000 recorded program edits.

\begin{table}[t]
\centering
\caption{Evolutionary coding frameworks in EvoTrace.}
\label{tab:frameworks}
\small
\setlength{\tabcolsep}{8pt}
\renewcommand{\arraystretch}{1.2}
\begin{tabular}{@{}lll@{}}
\toprule
\textbf{Framework} & \textbf{Search strategy} & \textbf{Distinctive feature} \\
\midrule
OpenEvolve   & Archive-based, MAP-Elites style & Open-source reference baseline \\
GEPA         & Reflective prompt evolution     & Natural-language lessons drive mutation \\
EvoX         & Meta-evolution                  & Adapts the search strategy during the run \\
ShinkaEvolve & Sample-efficient bandit search  & Parent sampling, novelty rejection, model routing \\
\bottomrule
\end{tabular}
\end{table}

\begin{table}[t]
\centering
\caption{Models used in EvoTrace.}
\label{tab:models}
\small
\setlength{\tabcolsep}{8pt}
\renewcommand{\arraystretch}{1.2}
\begin{tabular}{@{}ll@{}}
\toprule
\textbf{Model} & \textbf{Role} \\
\midrule
\texttt{deepseek-reasoner}      & Primary generator \\
\texttt{claude-sonnet-4-6}      & Frontier alternative \\
\texttt{claude-haiku-4-5}       & Small reasoning model \\
\texttt{gemini-3-flash-preview} & Cross-vendor alternative \\
\texttt{deepseek-chat}          & Non-reasoning baseline \\
\bottomrule
\end{tabular}
\end{table}

\subsection{Trace schema and design choices}
\label{sec:trace_representation}

EvoTrace normalizes each run, regardless of which backend produced it, into a unified JSONL schema. The schema covers run-level metadata, candidate programs (full source, byte-identical), evaluator outputs (raw execution logs, errors, timings, task-specific metrics), parent--child edges with operator labels, the prompts and contexts the LLM saw at generation time, and the replay environment needed to rerun selected candidates against their original evaluator.

Recording the full source rather than a score-only log is what enables literal extraction for the BO baseline (\S\ref{sec:tuning-gap}), the cycling classifier (\S\ref{sec:cycling}), and same-prompt replay (\S\ref{sec:replay}). Replayability is treated as a collection criterion: traces we cannot rerun against their original evaluator are excluded.

The schema supports two complementary uses: aggregate analysis (cross-framework comparison of population sizes, score progression, diversity, validity, lineage depth, best-so-far trajectories) and local reconstruction by EvoReplay (\S\ref{sec:evoreplay}). The full per-field schema is given in Appendix~\ref{appendix:trace-schema}.

\section{EvoReplay}
\label{sec:evoreplay}

EvoTrace records what happened during a run; \emph{EvoReplay} is the Python package we built on top of it (and on top of SkyDiscover \citep{skydiscover2026}) to ask why. By treating each candidate program as an executable artifact attached to its evaluator, parent context, and search position, EvoReplay turns a passive log into an experimental object: any point in the search graph can be re-executed, perturbed, retuned, or re-judged, and the outcome compared against what the original run produced.

This section describes the four capabilities EvoReplay provides. Each experimental section in the rest of the paper uses one of them, and the package is the common substrate that makes our cross-framework, cross-model results comparable.

\paragraph{(a) Static analysis of traces.}
EvoReplay normalizes runs from different backends into a common per-edit table (parent, child, prompt, score, and a unified diff between parent and child source), so that aggregate measurements are defined once and applied across frameworks. The static analyses in \S\ref{sec:static-analysis} (hyperparameter-literal counts, lineage depth, best-so-far trajectories) and the deterministic cycling classifier in \S\ref{sec:cycling} both operate on this normalized representation, with no framework-specific code path.

\paragraph{(b) LLM-as-judge annotation of edit types.}
The 9-edit-type taxonomy referenced in the abstract and \S\ref{sec:static-analysis} was developed by the authors through manual inspection of parent--child edits sampled across frameworks, languages, and models, grouping recurring patterns and iterating until no new categories emerged. EvoReplay's pipeline then applies this taxonomy at scale: for every parent--child diff, it requests a structured judgment from an LLM judge, returning a category for the edit and tags for the lines that drove the score change. The package handles batching, retries, schema validation, and caching, so the same trace can be re-annotated under different prompts or judge models without re-running the underlying search.

\paragraph{(c) Bayesian optimization to study hyperparameter tuning.}
EvoReplay implements the BO baseline of \S\ref{sec:tuning-gap}: a single LLM call identifies tunable numeric constants in a target program (full prompt in Appendix~\ref{appendix:bo-details}), the package rewrites the program with a top-level parameter block, and \texttt{gp\_minimize} runs the same evaluator harness with 24 calls per target. This isolates the structural-vs-parametric component of an evolutionary gain on a fixed seed structure, and lets us quantify how much of $f^{\star}_{\mathrm{evo}}$ a single-seed hyperparameter sweep already recovers.

\paragraph{(d) Stability analysis of breakthroughs.}
EvoReplay can re-execute the saved generating prompt for any candidate under the original or a substituted model and report the distribution over children. The replay-stability results of \S\ref{sec:replay} use this capability with $n{=}10$ resamples per target across same-model and cross-model conditions; the resulting (parse success) $\times$ (evaluation success) $\times$ (score conditional on success) triple is the right summary because failure modes turn out to be bimodal rather than Gaussian.

Together, these four capabilities make EvoTrace more than a collection of examples: they let us ask which improvements are reproducible, which are parametric, which are structural, and how each framework's edit composition shifts across models and prompting modes.

\section{Results}
\label{sec:experiments}

We analyze $121$ evolutionary runs across the four frameworks of Table~\ref{tab:frameworks}, $16$ tasks spanning Python mathematical constructions and C++ ALE-bench problems (\texttt{ahc008}--\texttt{ahc046}), and $5$ LLMs varied with and without diff-based generation (whether the LLM emits a unified diff or a full rewritten program). Each run consists of $100$ search iterations. Aggregate program-, call-, and token-level statistics are reported in Appendix~\ref{appendix:experiment-scale}.

\subsection{Static analysis: what gets evolved?}
\label{sec:static-analysis}

\begin{figure}[t]
\centering
\includegraphics[width=0.95\linewidth]{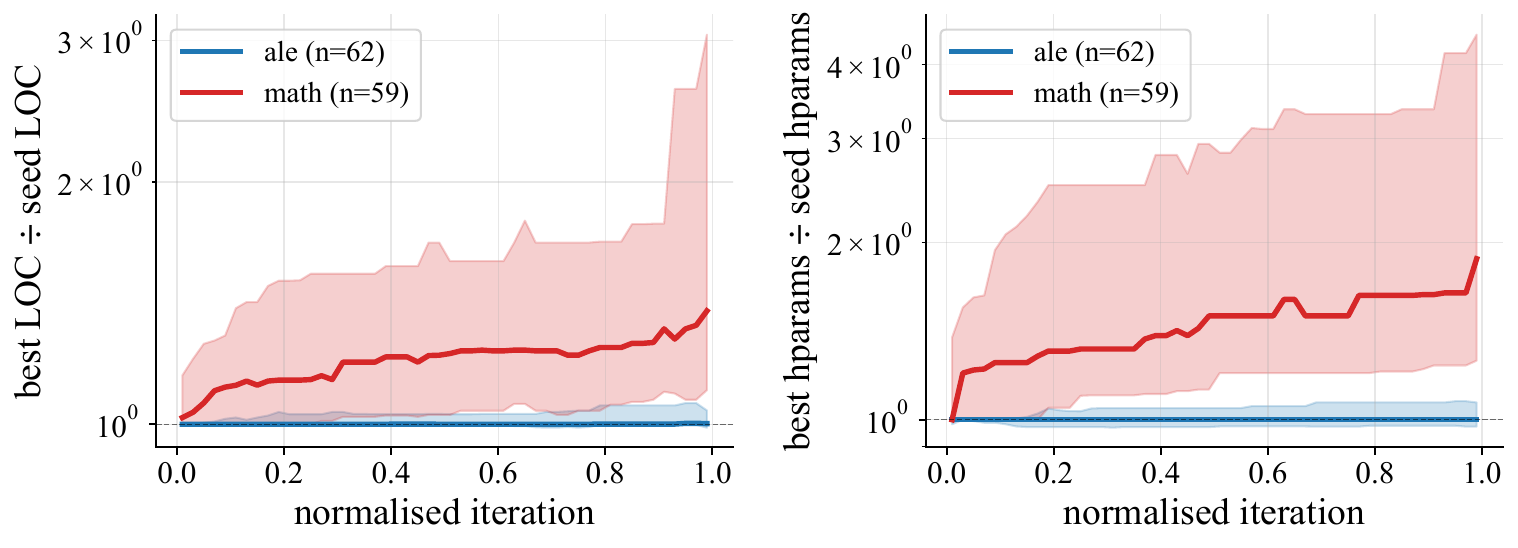}
\caption{\textbf{Program size and numeric-literal hyperparameter count over a run.} Best-so-far program length (LOC, left) and numeric-literal count (right), each normalized by the run's seed value, plotted against normalized iteration. Solid line = cross-run median; shaded band = inter-quartile range; dashed gray line marks the seed value. Math runs ($n{=}59$) accumulate modest LOC and hp growth (median final ratios $1.33\times$ and $1.70\times$); ALE runs ($n{=}62$) refine large seeds in place (final ratio $\approx 1.0\times$ on both axes).}
\label{fig:loc-hparam}
\end{figure}

We characterize the typical behaviour of evolutionary coding frameworks across our $121$ runs. Programs do change shape during search (Figure~\ref{fig:loc-hparam}): math runs accumulate modest LOC and numeric-literal growth, while ALE runs are refined at near-constant size from already-large seeds. The figure serves as backdrop for the math-comparison difficulty discussed next and for the BO-ceiling probe of \S\ref{sec:tuning-gap}; the more interesting question is what these changes contribute, which we address through edit-level analysis below.

\paragraph{Lineage depth and budget utilization.}
The chain of parents from each final-best program back to the seed is short in both domains, but is somewhat shorter on ALE (median lineage depth $4$, vs.\ $6$ on math). On math the median best-so-far is reached around iteration $0.75$ of the budget, while on ALE it lands earlier (median normalized iteration $0.49$). In both cases the dominant pattern is jackpot-then-flat: most of the per-run iteration budget is spent on dead branches that do not contribute to the final best.

\paragraph{Public-vs-private generalization on ALE.}
ALE-bench public scores are not the held-out judging metric. Re-scoring every ALE run's public best-so-far chain on the private test set used by AtCoder ($n{=}30$ run/problem pairs across the four main backends), two of the four frameworks overfit on at least $30\%$ of the problems they were scored on, and the same problem can flip generalization sign across frameworks: on \texttt{ahc024}, OpenEvolve found a $+1{,}606$ rating-point private gain while ShinkaEvolve, on the same problem, \emph{lost} $1{,}610$ rating points despite a positive public score change. The public best-so-far chain is therefore unreliable as a single-number summary on ALE; full per-problem and per-framework tables are in Appendix~\ref{appendix:public-private}.

\begin{figure}[t]
\centering
\begin{minipage}[b]{0.49\linewidth}
  \centering
  \includegraphics[width=\linewidth]{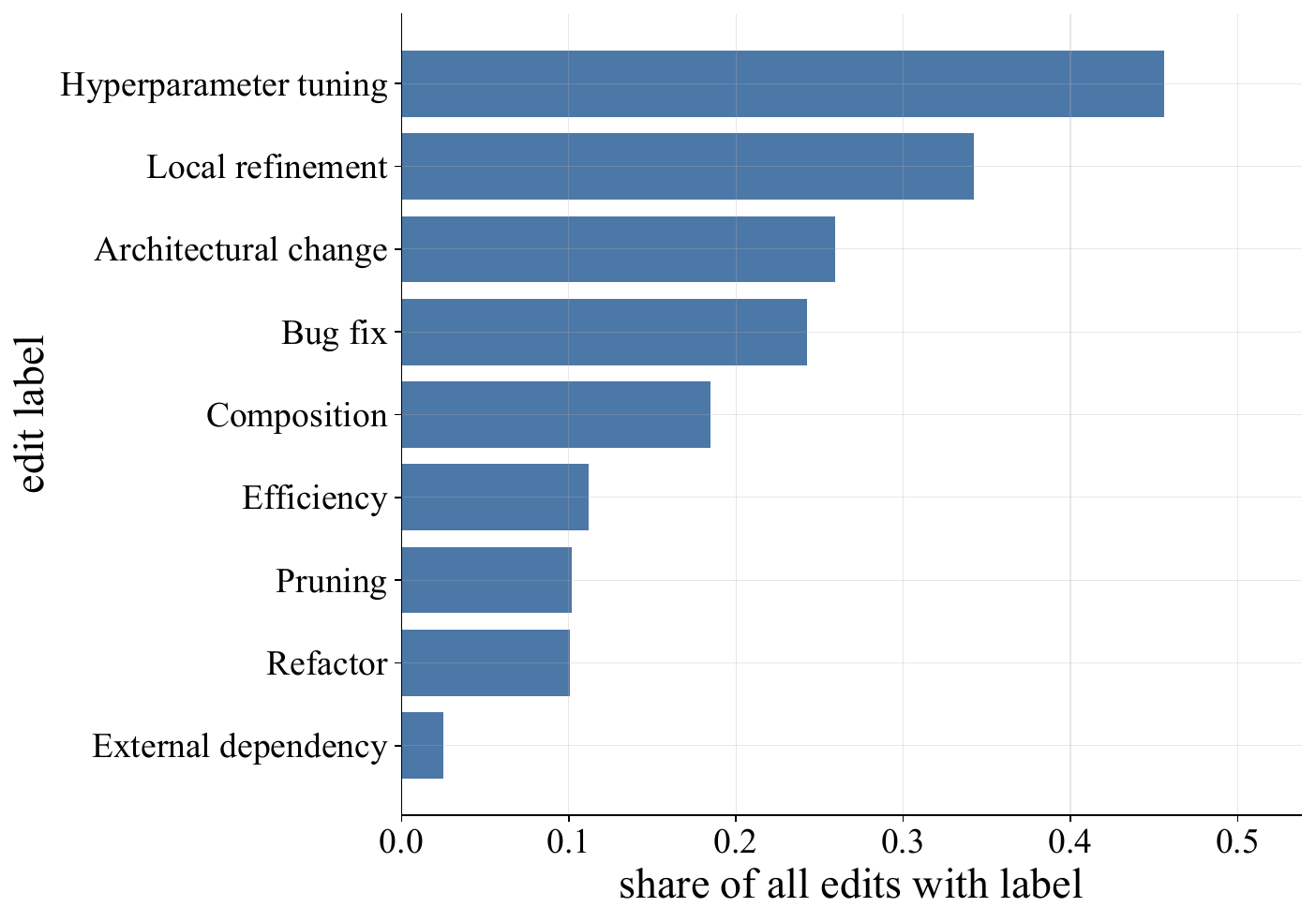}\\
  {\small (a) Prevalence}
\end{minipage}\hfill
\begin{minipage}[b]{0.49\linewidth}
  \centering
  \includegraphics[width=\linewidth]{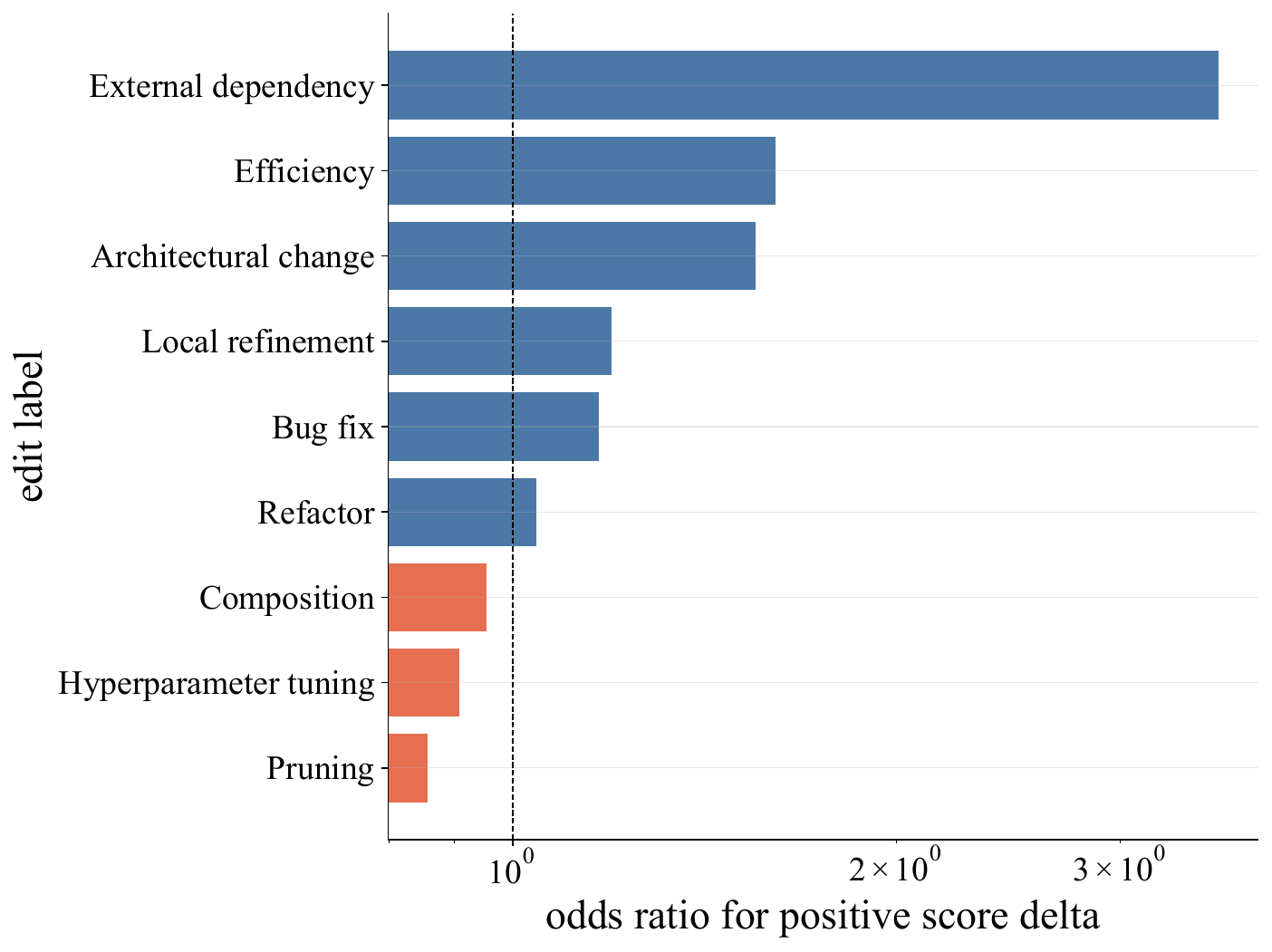}\\
  {\small (b) Helpfulness (odds ratio)}
\end{minipage}
\caption{\textbf{Edit-taxonomy: frequency vs.\ per-edit utility} across all programs in EvoTrace. \textbf{(a)}~Frequency of each label: \emph{Hyperparameter tuning} dominates the search distribution. \textbf{(b)}~Per-edit odds ratio for positive normalized score change: \emph{External dependency}, \emph{Efficiency}, and \emph{Architectural change} are the most helpful categories on a per-edit basis. The categories that most often improve a single edit are not the categories the search spends most of its effort on. Best-so-far and final-best-lineage enrichment views, plus per-domain and per-backend breakdowns, are in Appendix~\ref{appendix:edit-taxonomy-aggregate} and Appendix~\ref{appendix:edit-taxonomy-breakdowns}.}
\label{fig:edit-taxonomy-results}
\end{figure}

\paragraph{Edit taxonomy via LLM-as-judge.}
We use EvoReplay's LLM-as-judge pipeline (\S\ref{sec:evoreplay}, capability (b)) to annotate every parent--child edit with one or more categories from a 9-label taxonomy (\emph{Hyperparameter tuning}, \emph{Local refinement}, \emph{Architectural change}, \emph{Composition}, \emph{Efficiency}, \emph{Bug fix}, \emph{External dependency}, \emph{Pruning}, and \emph{Refactor}, applied to every parent--child edit in EvoTrace across the four backends. Agreement between the judge and a blind human re-annotation on a stratified sample of $200$ edits is substantial overall (macro $\kappa = 0.77$, micro-$F_1 = 0.90$, exact-match accuracy $74.5\%$), with per-category breakdowns and the one failure case (\emph{external\_dependency}) reported in Appendix~\ref{appendix:judge-validation}. The picture splits cleanly into a frequency view and a per-edit utility view (Figure~\ref{fig:edit-taxonomy-results}).
By frequency, \emph{Hyperparameter tuning} is the single most prevalent label, consistent with the cycling pattern of \S\ref{sec:cycling} and the tuning-gap analysis of \S\ref{sec:tuning-gap}. By per-edit utility, however, the strongest categories are different: \emph{External dependency} edits have a $3.58\times$ odds ratio for positive normalized score change ($n{=}104$), \emph{Efficiency} $1.61\times$ ($n{=}464$), and \emph{Architectural change} $1.55\times$ ($n{=}1{,}075$).
The frequency--utility gap propagates to successful trajectories: best-so-far updates and final-best lineages are both enriched in \emph{Efficiency}, \emph{External dependency}, and \emph{Hyperparameter tuning} relative to the all-edits base rate (Appendix~\ref{appendix:edit-taxonomy-aggregate}). Edits are typically multi-label ($67.4\%$ have $\geq 2$ labels), so these categories should not be read as mutually exclusive modes; the most common compound patterns are \emph{Hyperparameter tuning + Local refinement} and \emph{Composition + Hyperparameter tuning}.

\finding{The categories that most often improve a single edit (\emph{External dependency}, \emph{Efficiency}, \emph{Architectural change}) are not the categories evolutionary search spends most of its effort on (\emph{Hyperparameter tuning}, \emph{Local refinement}). Most score gains come from a small subset of edit types, and that subset is rare in the search distribution.}

\subsection{Cycling: re-introducing previously deleted code}
\label{sec:cycling}

While manually inspecting traces to derive the edit taxonomy, we repeatedly observed lineages re-introducing lines they had earlier deleted, so we operationalized this as a deterministic check: for each parent--child diff, how often is an \emph{added} line byte-identical to a line that the same lineage has already \emph{deleted} in an earlier iteration? Across all $121$ runs, the median share of added lines that are such re-introductions is $\sim\!30\%$, and this rate grows monotonically over the run in $118$ of $121$ cases (median per-iteration slope $+0.0030$). Cycling is present throughout the trajectory of essentially every run we measure, not a late-run pathology, and is dominated by short-span churn (median $5$ iterations between deletion and re-introduction); the signal is stable across all four frameworks, both languages, and all $5$ generator models we tested. A walk-through of one short-span cycle is given in Appendix~\ref{appendix:cycling-example}, with additional analyses (a finer three-way recycling classifier, a model- and prompt-dependence breakdown, and a null result on post-breakthrough cycling) in Appendix~\ref{appendix:cycling-extras}.

\finding{Roughly $30\%$ of code lines added during evolutionary search are byte-identical to lines previously deleted in the same lineage, and the cycling rate grows monotonically over the run in $118$ of $121$ cases. Search budget is partly spent re-introducing material the run has already discarded, a deterministic and reproducible signal that is present throughout each run and across all frameworks, languages, and generator models we tested.}

\subsection{Replay reproducibility: structural, not lexical}
\label{sec:replay}

When evolutionary search reaches a new best-so-far program, can we reproduce that breakthrough by re-running the same prompt? For each of $36$ best-so-far events across the four backends, we re-prompted an LLM $10$ times with the exact context the original run had used and asked four questions of each replayed program: does it run? does the evaluator accept it? does it match the original program byte-for-byte? does it match the original score? Table~\ref{tab:replay-summary} reports the medians and four illustrative targets covering the recurring patterns.

\finding{Replays almost always produce a runnable program (median parse and evaluator success $1.00$) but essentially never the original program (median exact-match $0.00$). They nevertheless recover a median $0.76$ of the original score from a \emph{different} program: the score gain is broadly reproducible from the same prompt context even though the specific program is not.}

\begin{table}[t]
\centering
\caption{\textbf{Replay summary across $36$ breakthrough events.} Aggregate medians and four illustrative targets covering recurring replay patterns. ``Replay/Original'' is the median replayed score divided by the original program's score.}
\label{tab:replay-summary}
\small
\setlength{\tabcolsep}{4pt}
\renewcommand{\arraystretch}{1.15}
\begin{tabular}{@{}lccccl@{}}
\toprule
Target & Parse & Eval & Exact & Replay/Orig.\ & Pattern \\
\midrule
\textbf{Median across $36$ targets} & $1.00$ & $1.00$ & $0.00$ & $0.76$ & --- \\
\midrule
\texttt{circle\_packing\_iter68\_improve}    & $1.00$ & $1.00$ & $\geq\!0$ & $\approx\!1.00$ & tight reproduction \\
\texttt{second\_autocorr\_iter79\_improve}   & $1.00$ & $1.00$ & $0.00$    & $0.96$          & parent-revert \\
\texttt{circle\_packing\_iter32\_neutral}    & $0.50$ & $0.50$ & $0.00$    & ---             & bimodal failure \\
\texttt{heilbronn\_convex\_iter100\_regress} & $1.00$ & $1.00$ & $0.00$    & $1.11$          & replays exceed regress.\ \\
\bottomrule
\end{tabular}
\end{table}

\subsection{The tuning gap: how much is just hyperparameter search?}
\label{sec:tuning-gap}

\newcommand{\NBOwins}{22}
\newcommand{\NBOnochange}{8}
\newcommand{\NBOregress}{6}

We separate a program $p$ into a structure $s$ and a hyperparameter vector $\theta \in \Theta_s$ it exposes, writing $p = s(\theta)$. Holding $s_0$ fixed and running Bayesian optimization over $\theta$ yields a tuning ceiling $f^{\star}_{\mathrm{BO}}(s_0) = \max_{\theta} f(s_0(\theta))$, and the \emph{tuning gap} $\Delta(s_0) = f^{\star}_{\mathrm{evo}} - f^{\star}_{\mathrm{BO}}(s_0)$ measures how much of the evolutionary gain reflects structural discovery rather than parametric search. We operationalize $f^{\star}_{\mathrm{BO}}$ with one \texttt{deepseek-reasoner} call that proposes per-knob log/linear intervals, an automatic rewrite to a top-level \texttt{PARAMS} block, and a $24$-call \texttt{gp\_minimize} ($8$ random $+ 16$ BO acquisitions; full pipeline in Appendix~\ref{appendix:bo-details}).

\begin{wraptable}{r}{0.42\textwidth}
\vspace{-1.0em}
\centering
\caption{BO outcomes on $36$ mid-run programs (median $6$ knobs per program; $24$ evaluator calls each).}
\label{tab:bo-outcome}
\small
\setlength{\tabcolsep}{8pt}
\renewcommand{\arraystretch}{1.15}
\begin{tabular}{@{}lr@{}}
\toprule
\textbf{Outcome} & \textbf{\# targets} \\
\midrule
BO improves over original & \NBOwins{} \\
No change                 & \NBOnochange{} \\
BO regresses              & \NBOregress{} \\
\midrule
\textbf{Total}            & \textbf{36} \\
\bottomrule
\end{tabular}
\vspace{-0.6em}
\end{wraptable}

\paragraph{BO matches the evolutionary run's final-best on most intermediate programs.}
On $36$ intermediate programs sampled across runs, frameworks, and models, BO improves over the program's original score in \NBOwins{} of $36$ cases (Table~\ref{tab:bo-outcome}). When compared against the run's \emph{final}-best score (rather than the program's original score), BO matches or exceeds it on $13$ of $15$ intermediate programs (median delta $+0.025$). The largest individual gain is on \texttt{heilbronn\_tri\_dsr\_nodiff}, where the evo run reached $0.521$ in $100$ iterations and BO on an intermediate program from the same run reached $0.886$ ($1.70\times$ the evo final-best). The strong dependence of $f^{\star}_{\mathrm{BO}}(s_0)$ on $s_0$'s exposed knobs complicates math-benchmark cross-framework comparison: two frameworks with similar topology but different knob exposures give different headline scores even with identical search behaviour, so the defensible per-target summary is the pair $\big(f(p_0),\, f^{\star}_{\mathrm{BO}}(s_0)\big)$.

\finding{A $24$-call Bayesian-optimization pass on a single intermediate program's exposed hyperparameters improves over the program's score in \NBOwins{} of $36$ probed targets, and matches or exceeds the evolutionary run's final-best score on $13$ of $15$ intermediate programs (median delta $+0.025$). On these targets, late evolutionary iterations on math are largely matched by post-hoc hyperparameter tuning of an earlier program.}

\section{Discussion}
\label{sec:discussion}

Looking at the traces themselves rather than at final scores, our diagnostics surface several recurring inefficiencies in current LLM-driven evolutionary code search. A non-trivial share of the search budget is spent re-introducing material the run has already discarded: $\sim\!30\%$ of added lines are byte-identical to previously-deleted ones, and this share grows steadily across the trajectory in $118$ of $121$ runs. Breakthrough events are also not crisp, repeatable artifacts: same-prompt replays almost never reproduce the original program byte-for-byte, yet typically recover a substantial fraction of its score from a \emph{different} program. The trajectory carries the structural gain, while the specific program is one draw from a wider distribution. Lineages back to the seed are short, so most of the per-run budget is spent on branches that do not contribute to the final best. On math benchmarks, a Bayesian-optimization pass over a single intermediate program's exposed knobs often matches or exceeds the run's final-best score, suggesting that on math the parametric refinement evolutionary search performs late in a run is largely substitutable by post-hoc tuning. On ALE, two of four frameworks overfit on at least $30\%$ of their problems. These patterns hold across four frameworks, two languages, and five LLMs.

\paragraph{Implications.}
On math, the BO finding suggests a natural decomposition of the work an evolutionary run does: the structural changes it makes early in a run, and the parametric refinement of those structures, which can often be done post-hoc by hyperparameter tuning of an intermediate program. A practical consequence is that math-benchmark headline scores should be reported alongside the single-program tuning ceiling $f^{\star}_{\mathrm{BO}}(s_0)$ so this decomposition is visible; on ALE, public scores should additionally be paired with a private-test re-score to surface overfitting. For system design, cycling growth and lineage shallowness suggest that interventions which prevent the search from re-doing discarded work (lineage-aware credit assignment, deletion-aware novelty filters, prompting strategies that expose a parent's deletion history) are promising directions to explore, complementary to extending the search budget.

\paragraph{Scope and open questions.}
The trace-level view we develop here opens several natural directions. The single-program tuning ceiling, currently reported on math, can be extended to other domains and evaluator types; replay-based reproducibility can be probed with larger samples and richer perturbations of the local search state; and the dynamics we surface (cycling, lineage shallowness, and the frequency--utility gap across edit categories) can be re-examined under different selection rules, prompting strategies, and underlying model families. Because EvoTrace records full source and replay environments, each of these follow-ups can be posed as a controlled intervention on the same trace, without re-running the original search.

\section*{Acknowledgements}
This research was partially supported by the DFG Cluster of Excellence MATH+ (EXC-2046/1, project id 390685689) funded by the Deutsche Forschungsgemeinschaft (DFG) as well as by the German Federal Ministry of Research, Technology and Space (fund number 16IS23025B).

\newpage

\bibliographystyle{unsrtnat}
\bibliography{references}

\newpage
\appendix

\section{Additional EvoTrace Details}
\label{appendix:evotrace-details}

\subsection{Per-field trace schema}
\label{appendix:trace-schema}

EvoTrace normalizes each run into six object types stored as JSONL tables, each motivated by a mechanistic question that storing only iteration-vs-score traces would foreclose. Table~\ref{tab:trace-schema} summarises the six object types.

\begin{table}[h]
\centering
\caption{EvoTrace per-field schema. Each object type is recorded because at least one analysis in the main paper requires the corresponding raw artifact rather than a score-only summary.}
\label{tab:trace-schema}
\small
\setlength{\tabcolsep}{6pt}
\renewcommand{\arraystretch}{1.25}
\begin{tabular}{@{}l p{4.6cm} p{6.0cm}@{}}
\toprule
\textbf{Object} & \textbf{Fields} & \textbf{Why recorded} \\
\midrule
Runs &
Task, framework, model configuration, evaluator command, seed artifact, search budget, run-time environment. &
Makes any reported number sliceable by framework or model. \\
Candidates &
Full program source (byte-identical), iteration index, parent identifiers, prompt context, validity status, evaluator score. &
The full source enables literal extraction for the BO baseline (\S\ref{sec:tuning-gap}, Appendix~\ref{appendix:bo-details}), the cycling classifier (\S\ref{sec:cycling}), and post-hoc simplification or repair. \\
Evaluations &
Execution outputs, error messages, timing, correctness signals, task-specific metrics. &
The bimodal failure picture in \S\ref{sec:replay} is invisible to score-only logging. \\
Edges &
Parent--child relations with the operator that produced the child (mutation, recombination, refinement, repair). &
Lineage-depth and dead-branch statistics in \S\ref{sec:static-analysis} require the full edge table, not the best-so-far trajectory. \\
Contexts &
Prompts, retrieved examples, population summaries, and lineage information seen by the LLM at generation time (byte-identical). &
Stability replays in \S\ref{sec:replay} pass the saved input back to the original or a substituted model. \\
Replay environments &
Evaluator command, dependencies, timeouts, hardware assumptions, raw artifacts. &
Replayability is a collection criterion: traces we cannot rerun against the original evaluator are excluded. \\
\bottomrule
\end{tabular}
\end{table}

\section{Additional Experimental Details}
\label{appendix:experimental-details}

\subsection{Experiment scale and cost}
\label{appendix:experiment-scale}

The dataset spans $121$ evolutionary runs that together propose $10{,}672$ unique programs (including $1{,}708$ explicitly rejected ones), make $18{,}400$ LLM calls, and consume $274.7$\,M prompt and $80.3$\,M completion tokens (of which $42.8$\,M are reasoning tokens). A typical $100$-iteration run produces about $100$ programs, makes about $134$ LLM calls, and uses $\sim 1.7$\,M prompt and $\sim 535$\,K completion tokens. ALE runs use approximately $2.8\times$ more prompt tokens than math runs because of their much larger seeds. Tables~\ref{tab:scale-by-backend}, \ref{tab:scale-by-domain}, and \ref{tab:scale-per-run} report full breakdowns.

\begin{table}[h]
\centering
\caption{Experiment scale by backend. ``Edits'' counts programs with a non-null parent; LLM calls and tokens are aggregated from each run's call log.}
\label{tab:scale-by-backend}
\small
\setlength{\tabcolsep}{6pt}
\renewcommand{\arraystretch}{1.15}
% Experiment scale by backend (paper appendix).
% Edits = programs with non-null parent_id; LLM calls and tokens
% are aggregates from each run's logs/llm_calls.jsonl.
\begin{tabular}{lrrrrrrrr}
\toprule
backend & runs & programs & accepted & rejected & edits & LLM calls & prompt tok & compl. tok \\
\midrule
openevolve & 44 & 4{,}267 & 4{,}267 & 0 & 4{,}223 & 5{,}064 & 84.9\,M & 21.1\,M \\
evox & 30 & 2{,}396 & 2{,}396 & 0 & 2{,}366 & 4{,}700 & 71.5\,M & 14.4\,M \\
gepa & 29 & 2{,}137 & 429 & 1{,}708 & 2{,}108 & 4{,}194 & 78.1\,M & 19.7\,M \\
shinka & 18 & 1{,}872 & 1{,}872 & 0 & 1{,}782 & 4{,}442 & 40.2\,M & 25.1\,M \\
\midrule
\textbf{total} & \textbf{121} & \textbf{10{,}672} & \textbf{8{,}964} & \textbf{1{,}708} & \textbf{10{,}479} & \textbf{18{,}400} & \textbf{274.7\,M} & \textbf{80.3\,M} \\
\bottomrule
\end{tabular}

\end{table}

\begin{table}[h]
\centering
\caption{Experiment scale by domain (ALE vs.\ math).}
\label{tab:scale-by-domain}
\small
\setlength{\tabcolsep}{6pt}
\renewcommand{\arraystretch}{1.15}
% Experiment scale by domain (ALE vs math). Excludes
% shinkaevolve\_private\_view (held-out re-evaluation only).
\begin{tabular}{lrrrrrr}
\toprule
domain & runs & programs & edits & LLM calls & prompt tok & compl. tok \\
\midrule
ale & 56 & 5{,}269 & 5{,}173 & 9{,}341 & 201.7\,M & 42.8\,M \\
math & 65 & 5{,}403 & 5{,}306 & 9{,}059 & 73.0\,M & 37.5\,M \\
\midrule
\textbf{total} & \textbf{121} & \textbf{10{,}672} & \textbf{10{,}479} & \textbf{18{,}400} & \textbf{274.7\,M} & \textbf{80.3\,M} \\
\bottomrule
\end{tabular}

\end{table}

\begin{table}[h]
\centering
\caption{Per-run cost (medians and right tails) over the $121$ runs.}
\label{tab:scale-per-run}
\small
\setlength{\tabcolsep}{8pt}
\renewcommand{\arraystretch}{1.15}
% Per-run cost (medians and right tails) for the 121 runs
% that actually called an LLM (excludes private\_view).
\begin{tabular}{lrrr}
\toprule
per-run metric & median & 75th pct & max \\
\midrule
programs                   & 101 & 107 & 118 \\
edits (parent\,\(\rightarrow\)\,child) & 99 & 106 & 117 \\
LLM calls                  & 134 & 165 & 512 \\
prompt tokens              & 1.7\,M & 3.3\,M & 7.3\,M \\
completion tokens          & 535{,}582 & 811{,}225 & 2.7\,M \\
\bottomrule
\end{tabular}

\end{table}

\subsection{Edit-taxonomy: aggregate enrichment views}
\label{appendix:edit-taxonomy-aggregate}

\begin{figure}[h]
\centering
\includegraphics[width=0.6\linewidth]{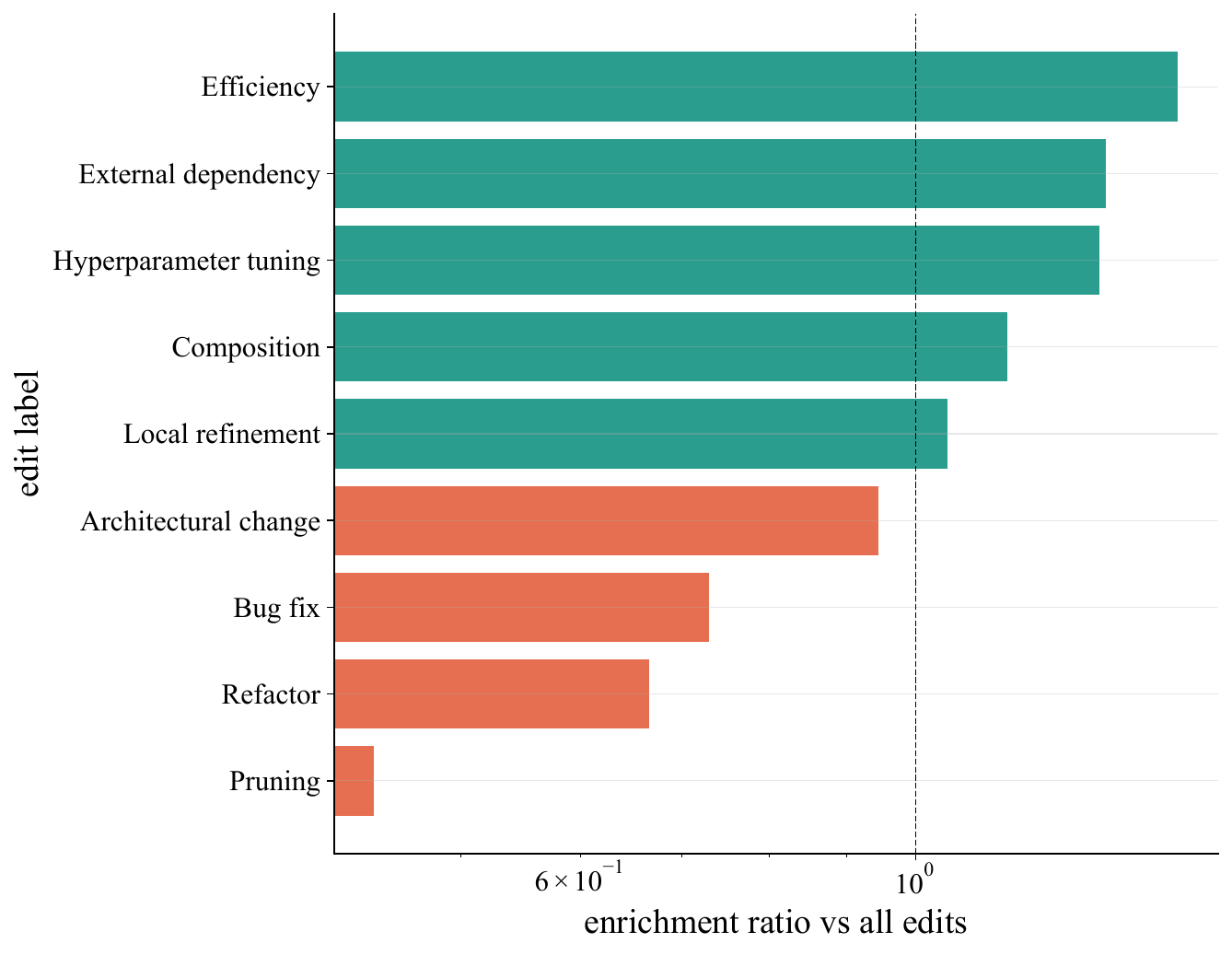}
\caption{\textbf{Best-so-far enrichment of edit labels (aggregate).} Enrichment of each taxonomy label among best-so-far updates relative to the all-edits base rate. The categories most overrepresented on successful intermediate steps (\emph{Efficiency}, \emph{External dependency}, \emph{Hyperparameter tuning}, \emph{Composition}) are not identical to the most frequent labels in Figure~\ref{fig:edit-taxonomy-results}~(a).}
\label{fig:edit-bsf-enrichment-aggregate}
\end{figure}

\begin{figure}[h]
\centering
\includegraphics[width=0.6\linewidth]{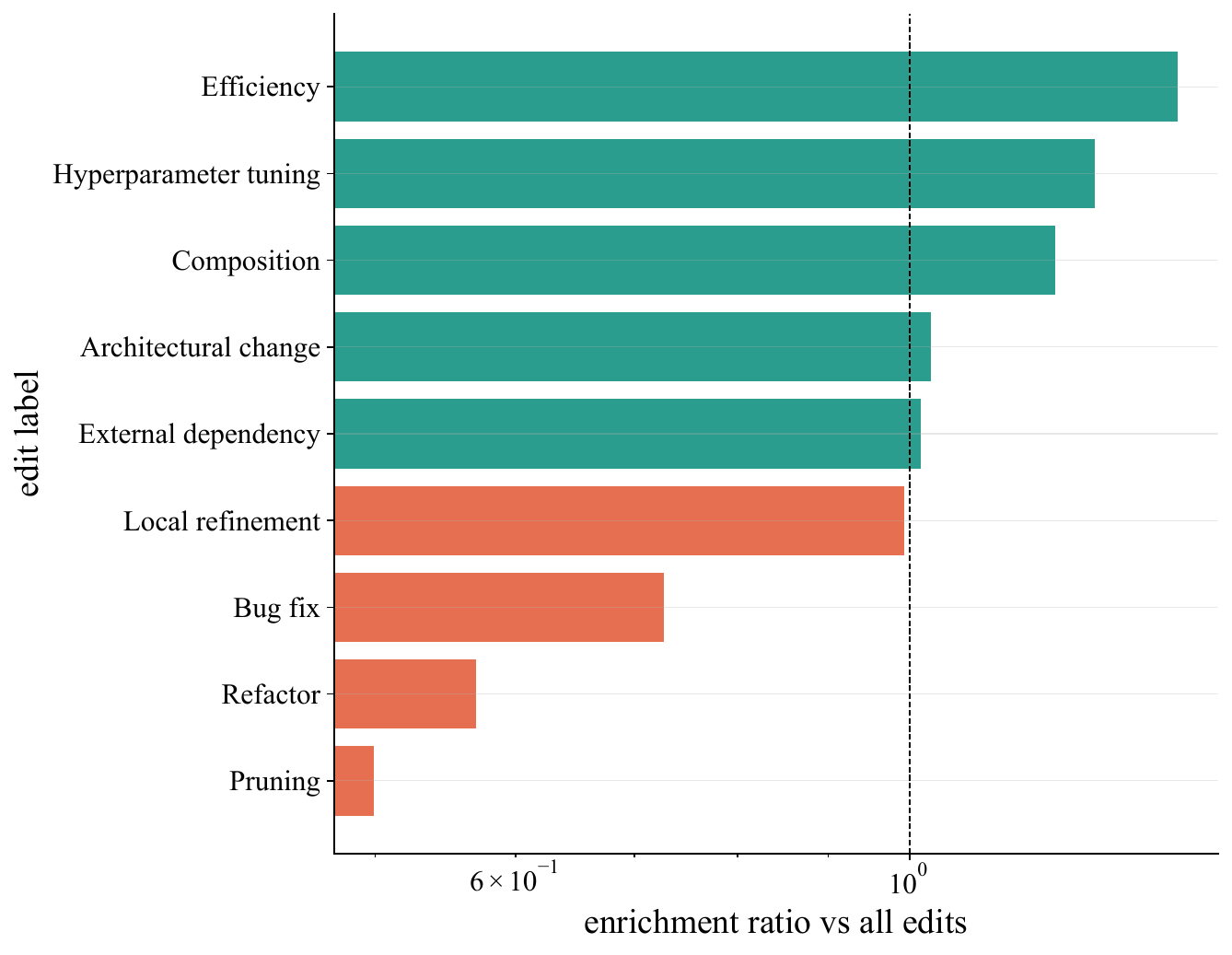}
\caption{\textbf{Final-best-lineage enrichment (aggregate, robustness check).} Enrichment of each label along the lineage from each run's final best program back to the seed. \emph{Efficiency}, \emph{Hyperparameter tuning}, and \emph{Composition} remain overrepresented relative to the all-edits base rate, supporting the best-so-far view in Figure~\ref{fig:edit-bsf-enrichment-aggregate}.}
\label{fig:edit-final-best-enrichment-aggregate}
\end{figure}

The main paper (Figure~\ref{fig:edit-taxonomy-results}) features the two-panel \emph{prevalence vs.\ helpfulness} view. The companion enrichment panels are reported here. Relative to the all-edits base rate, best-so-far updates are enriched in \emph{Efficiency} ($1.49\times$), \emph{External dependency} ($1.34\times$), and \emph{Hyperparameter tuning} ($1.32\times$); final-best lineages retain a similar mix, with \emph{Efficiency} ($1.42\times$), \emph{Hyperparameter tuning} ($1.27\times$), and \emph{Composition} ($1.21\times$) overrepresented (Figures~\ref{fig:edit-bsf-enrichment-aggregate} and \ref{fig:edit-final-best-enrichment-aggregate}). The same broad set of categories is enriched on both intermediate-improvement events and on the lineages that produce the eventual winner, so the frequency--utility gap surfaced in the main figure is not an artefact of conditioning on a particular subset of edits.

\subsection{Edit-taxonomy breakdowns by domain and backend}
\label{appendix:edit-taxonomy-breakdowns}

The aggregate edit-taxonomy results reported in \S\ref{sec:static-analysis} (Figure~\ref{fig:edit-taxonomy-results}) combine ALE and math runs across all four backends. This section reports the same analyses split by domain and, for the helpfulness view, by backend, so that readers can verify that the headline patterns survive these slices. The underlying labeled corpus covers all programs in EvoTrace.

\begin{figure}[h]
\centering
\begin{minipage}[b]{0.49\linewidth}
  \centering
  \includegraphics[width=\linewidth]{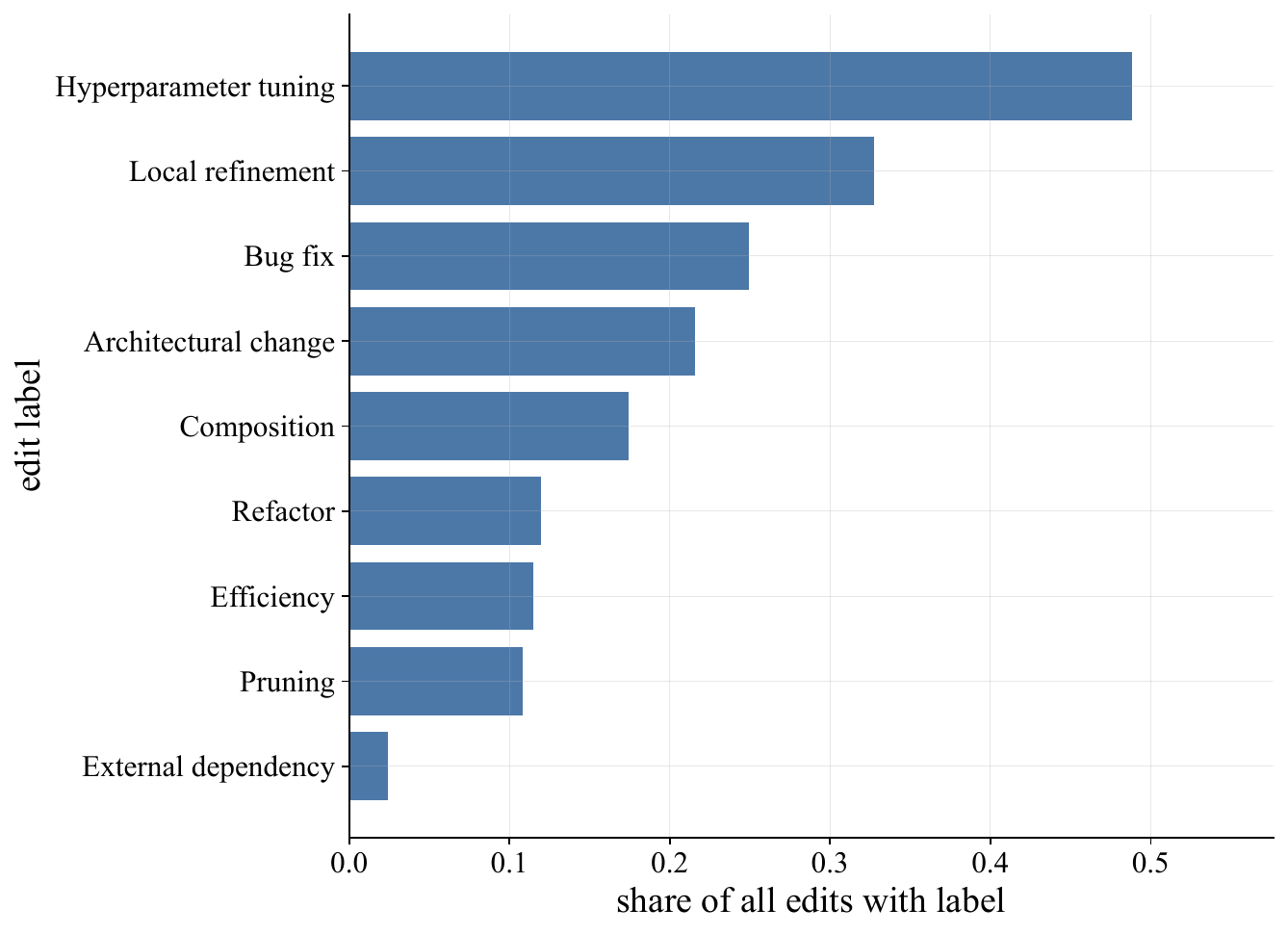}\\
  {\small (a) ALE}
\end{minipage}\hfill
\begin{minipage}[b]{0.49\linewidth}
  \centering
  \includegraphics[width=\linewidth]{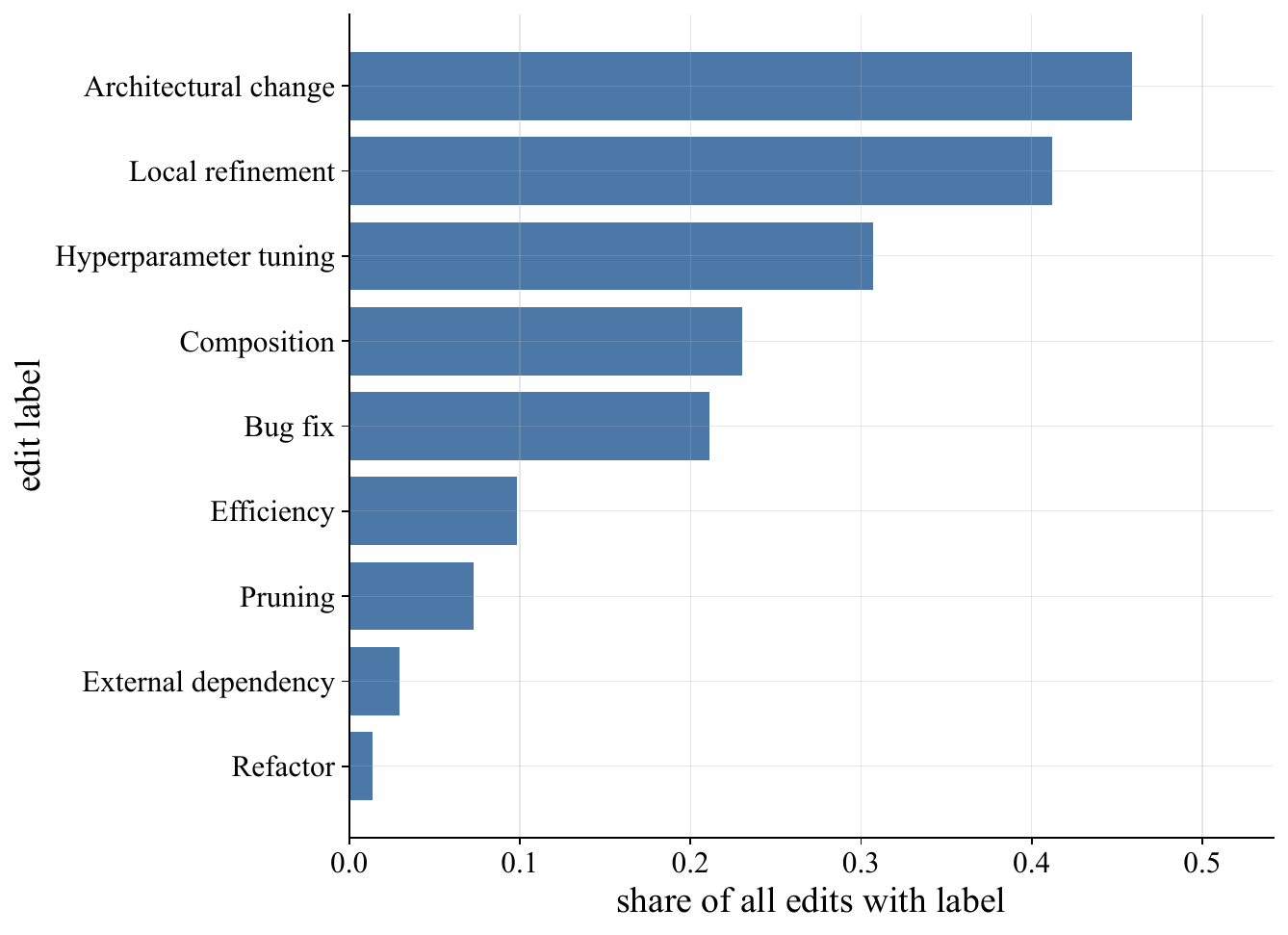}\\
  {\small (b) Math}
\end{minipage}
\caption{\textbf{Edit-label prevalence by domain.} Frequency of each taxonomy label among labeled edits, split by domain. \emph{Hyperparameter tuning} dominates in both domains, but \emph{Composition} is more prominent on math while structural categories shift their relative weights between domains.}
\label{fig:edit-prevalence-by-domain}
\end{figure}

\begin{figure}[h]
\centering
\begin{minipage}[b]{0.49\linewidth}
  \centering
  \includegraphics[width=\linewidth]{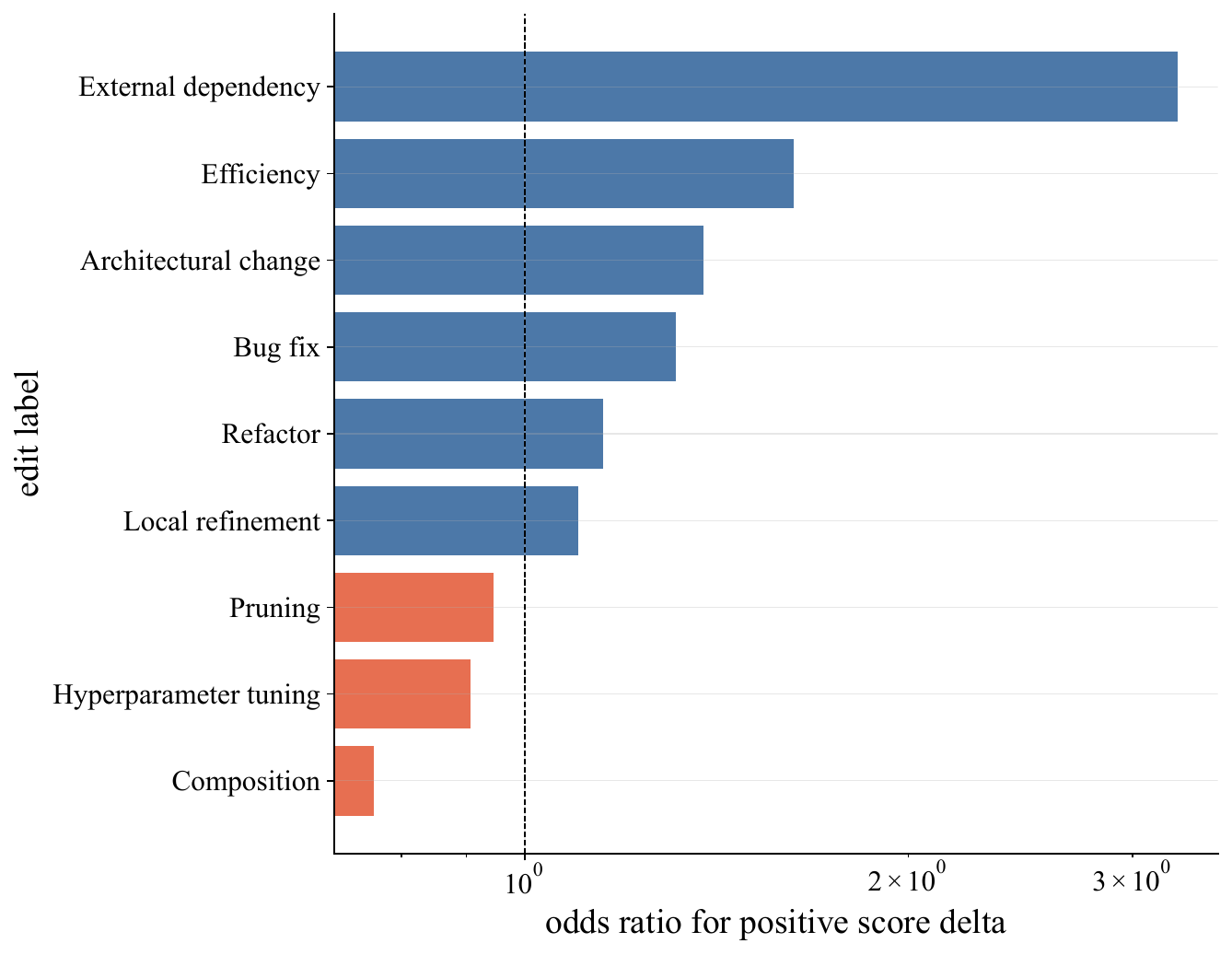}\\
  {\small (a) ALE}
\end{minipage}\hfill
\begin{minipage}[b]{0.49\linewidth}
  \centering
  \includegraphics[width=\linewidth]{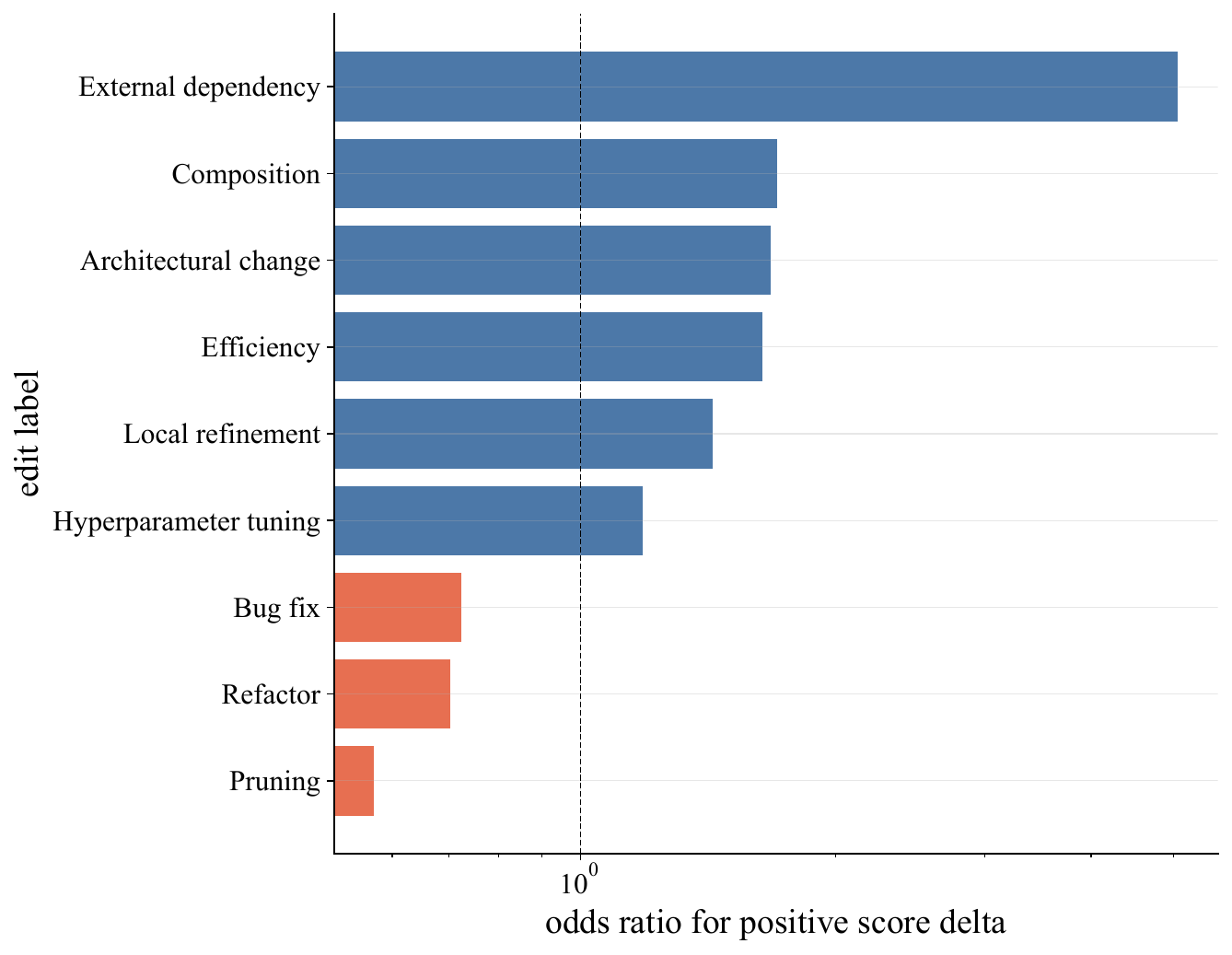}\\
  {\small (b) Math}
\end{minipage}
\caption{\textbf{Per-edit helpfulness (odds ratio for positive normalized score change) by domain.} On ALE, \emph{External dependency}, \emph{Efficiency}, and \emph{Architectural change} are the strongest positive categories. On math, \emph{External dependency} is even stronger and \emph{Composition} plays a larger role than on ALE.}
\label{fig:edit-helpfulness-by-domain}
\end{figure}

\begin{figure}[h]
\centering
\begin{minipage}[b]{0.49\linewidth}
  \centering
  \includegraphics[width=\linewidth]{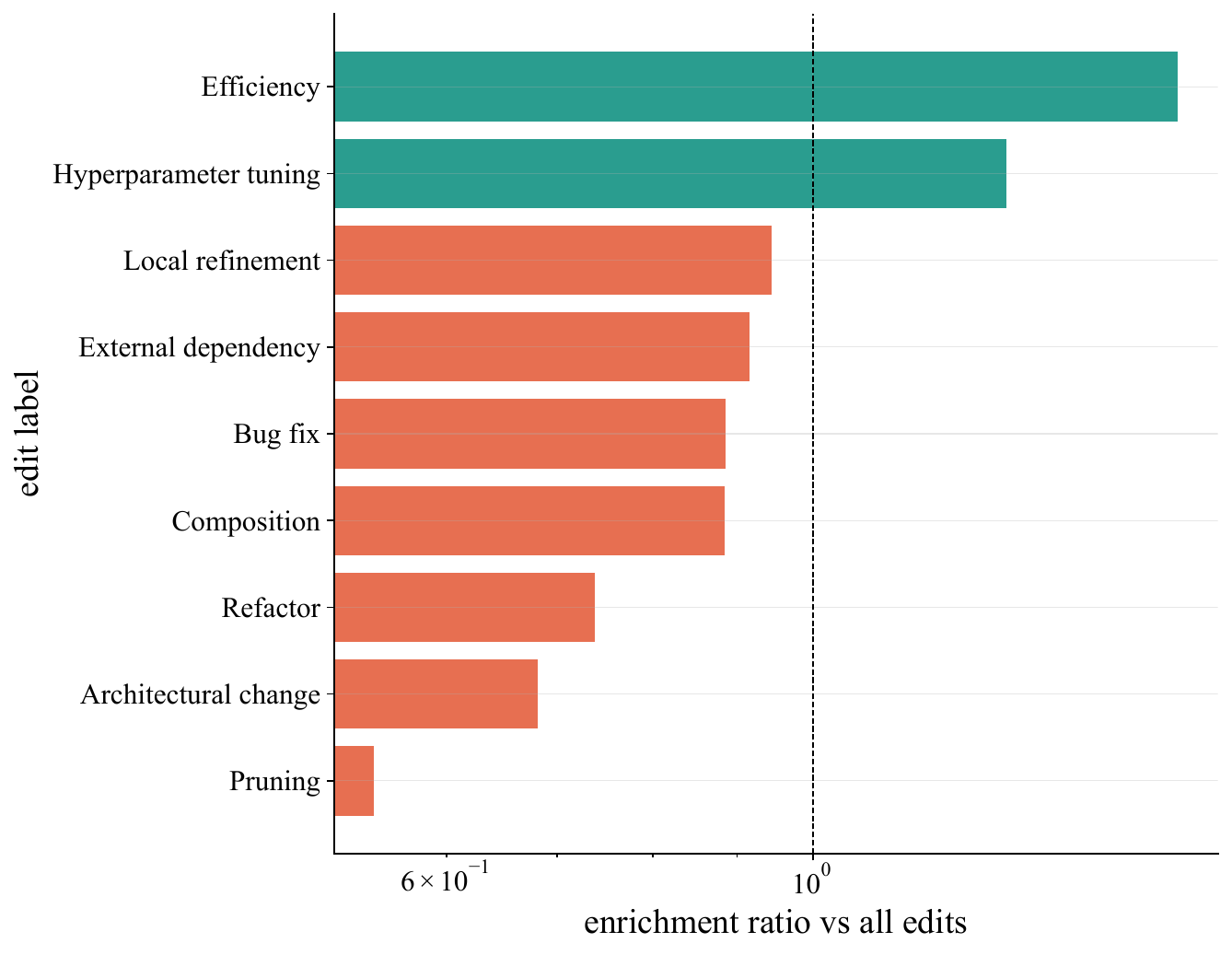}\\
  {\small (a) ALE}
\end{minipage}\hfill
\begin{minipage}[b]{0.49\linewidth}
  \centering
  \includegraphics[width=\linewidth]{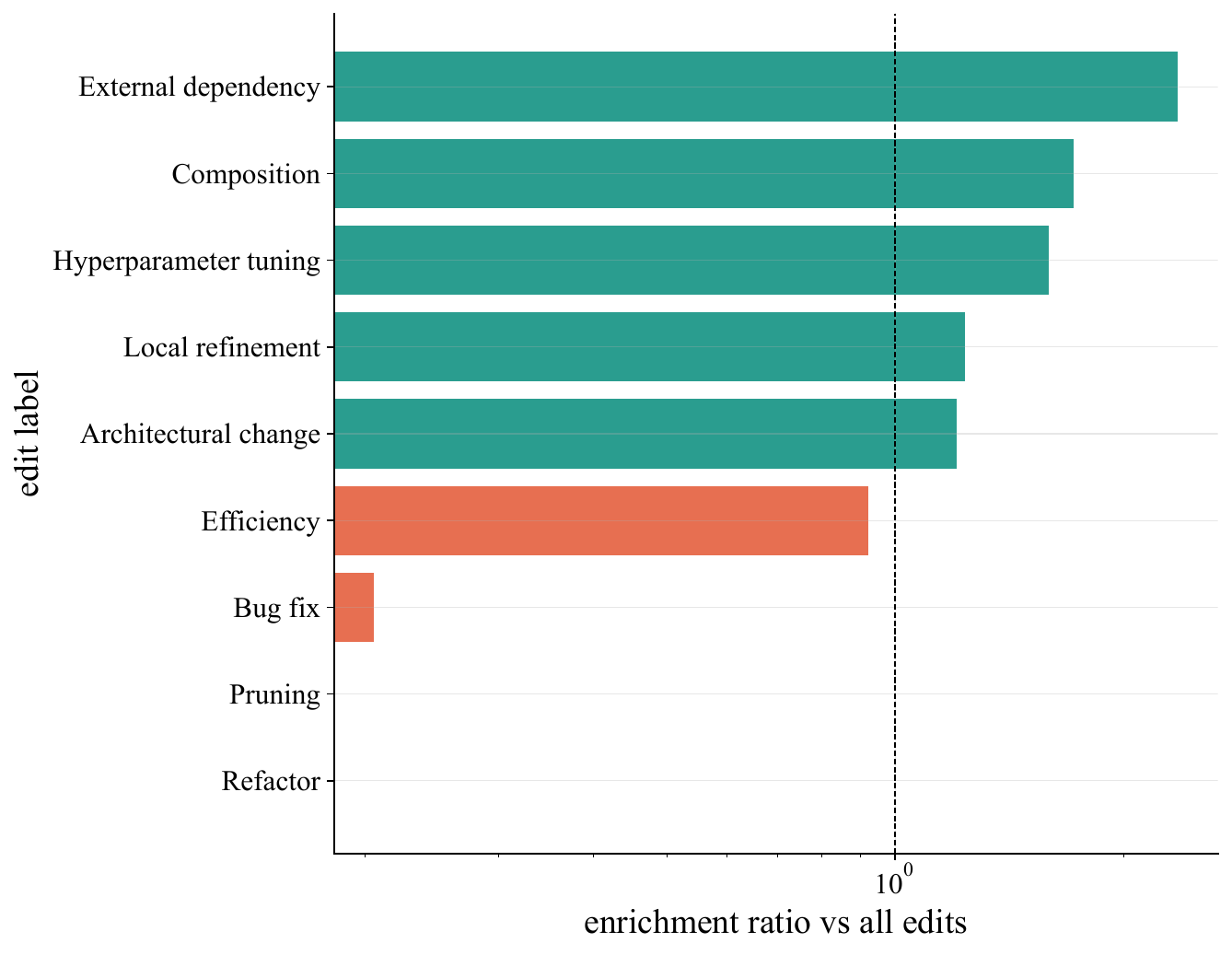}\\
  {\small (b) Math}
\end{minipage}
\caption{\textbf{Best-so-far enrichment of edit labels by domain.} Enrichment of each label among best-so-far updates relative to the all-edits base rate, split by domain. The qualitative signal, a small set of categories (notably \emph{Efficiency}, \emph{External dependency}, and \emph{Hyperparameter tuning}) overrepresented on successful intermediate steps, is consistent with the aggregate view in Figure~\ref{fig:edit-taxonomy-results}, with domain-specific shifts in magnitude.}
\label{fig:edit-bsf-enrichment-by-domain}
\end{figure}

\begin{figure}[h]
\centering
\begin{minipage}[b]{0.49\linewidth}
  \centering
  \includegraphics[width=\linewidth]{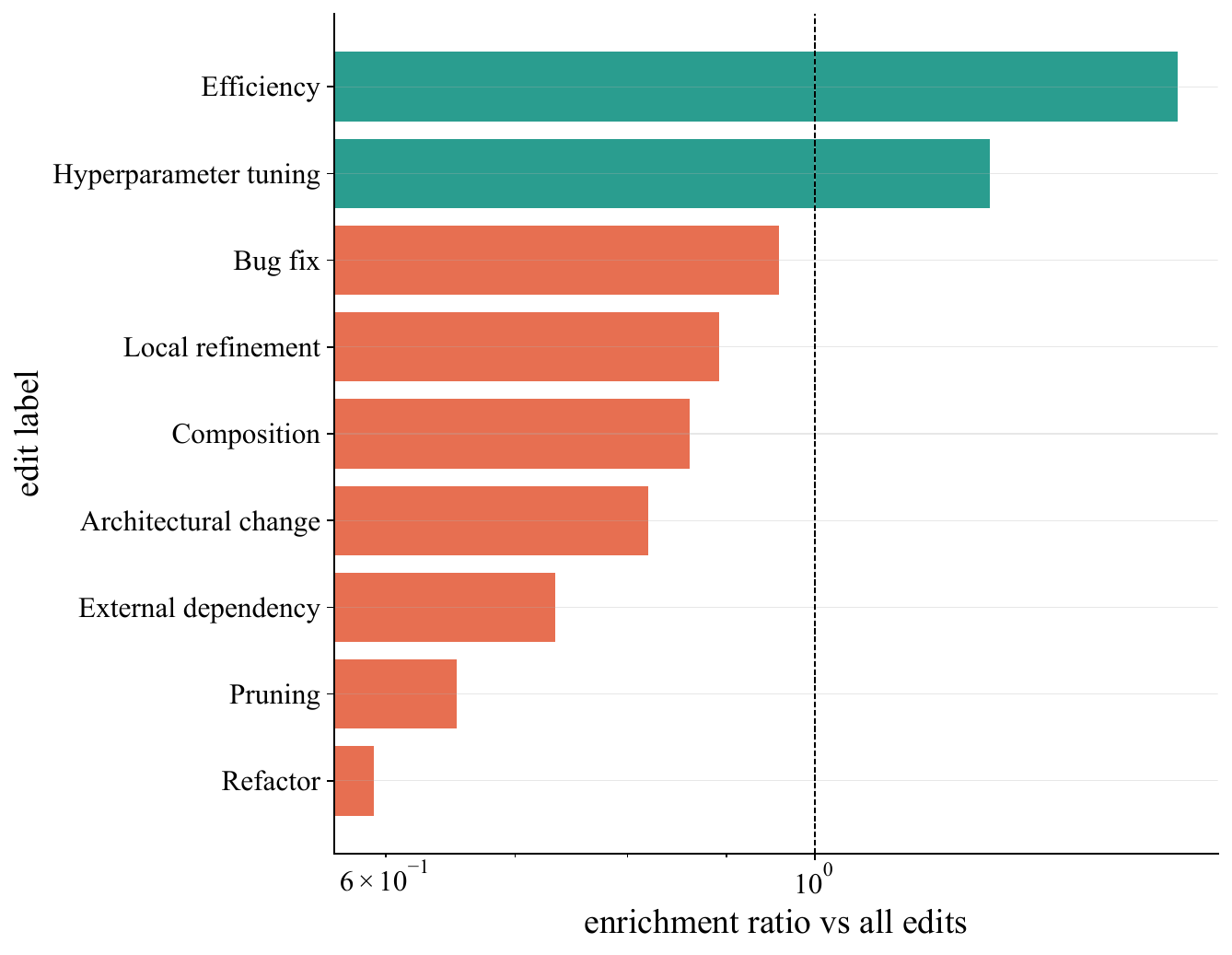}\\
  {\small (a) ALE}
\end{minipage}\hfill
\begin{minipage}[b]{0.49\linewidth}
  \centering
  \includegraphics[width=\linewidth]{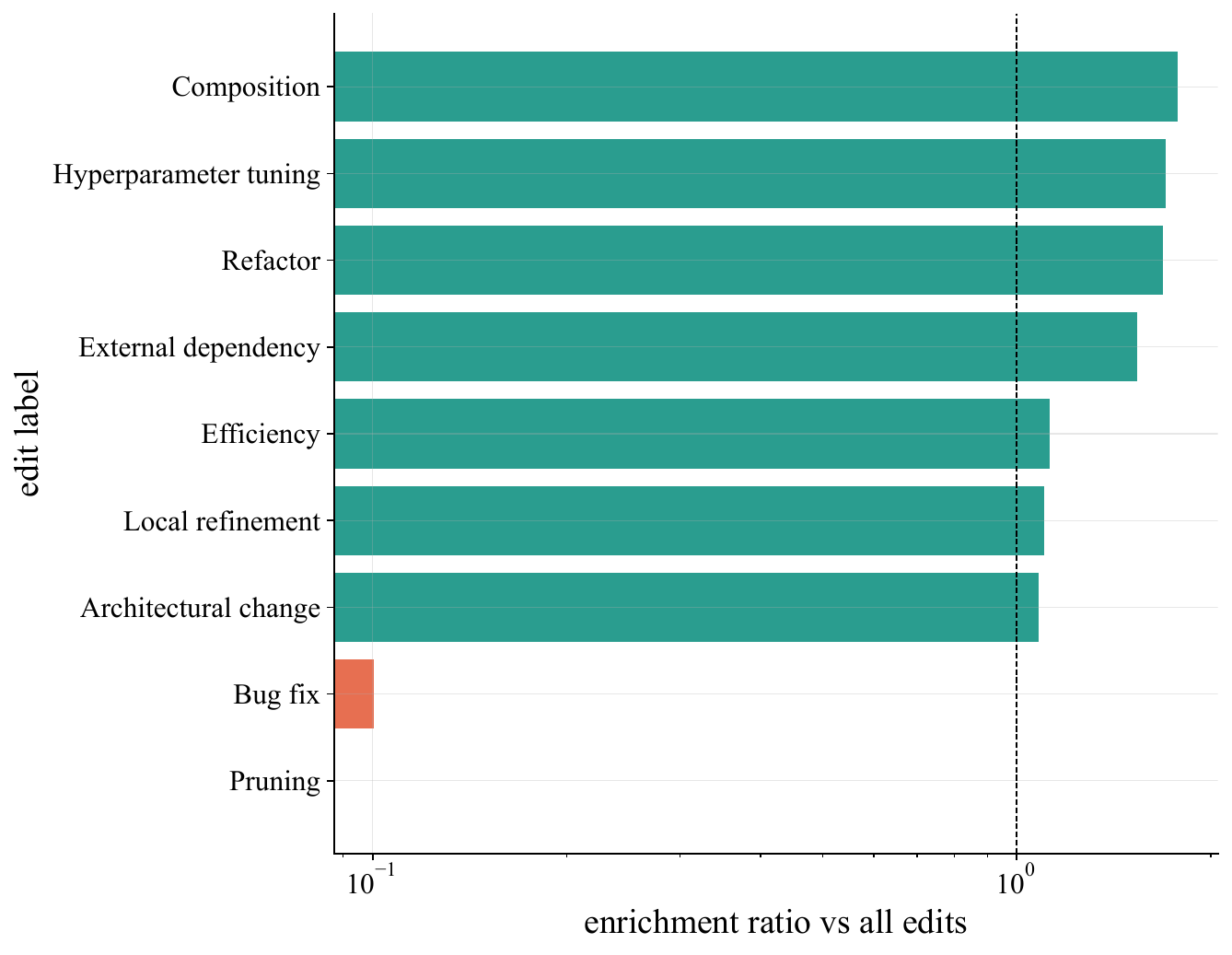}\\
  {\small (b) Math}
\end{minipage}
\caption{\textbf{Final-best-lineage enrichment by domain.} Robustness check for Figure~\ref{fig:edit-bsf-enrichment-by-domain}, restricted to edits that lie on the lineage of each run's final best program. The enriched categories overlap heavily with the best-so-far view in both domains, with \emph{Efficiency} and \emph{Hyperparameter tuning} retaining their overrepresentation.}
\label{fig:edit-final-best-enrichment-by-domain}
\end{figure}

\begin{figure}[h]
\centering
\begin{minipage}[b]{0.49\linewidth}
  \centering
  \includegraphics[width=\linewidth]{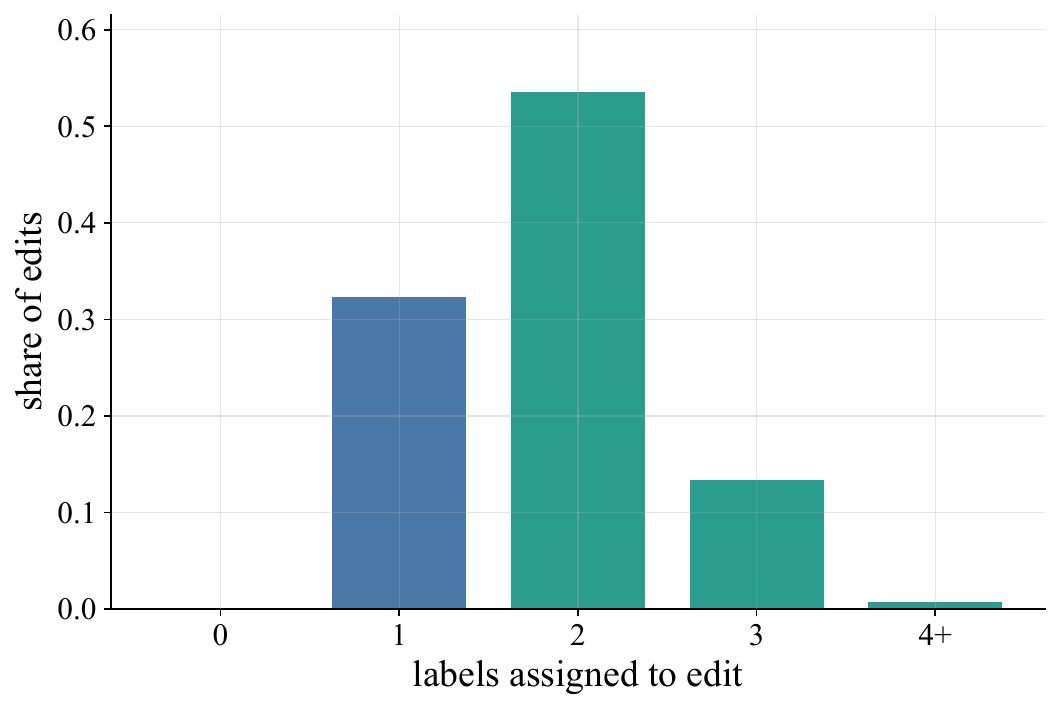}\\
  {\small (a) ALE}
\end{minipage}\hfill
\begin{minipage}[b]{0.49\linewidth}
  \centering
  \includegraphics[width=\linewidth]{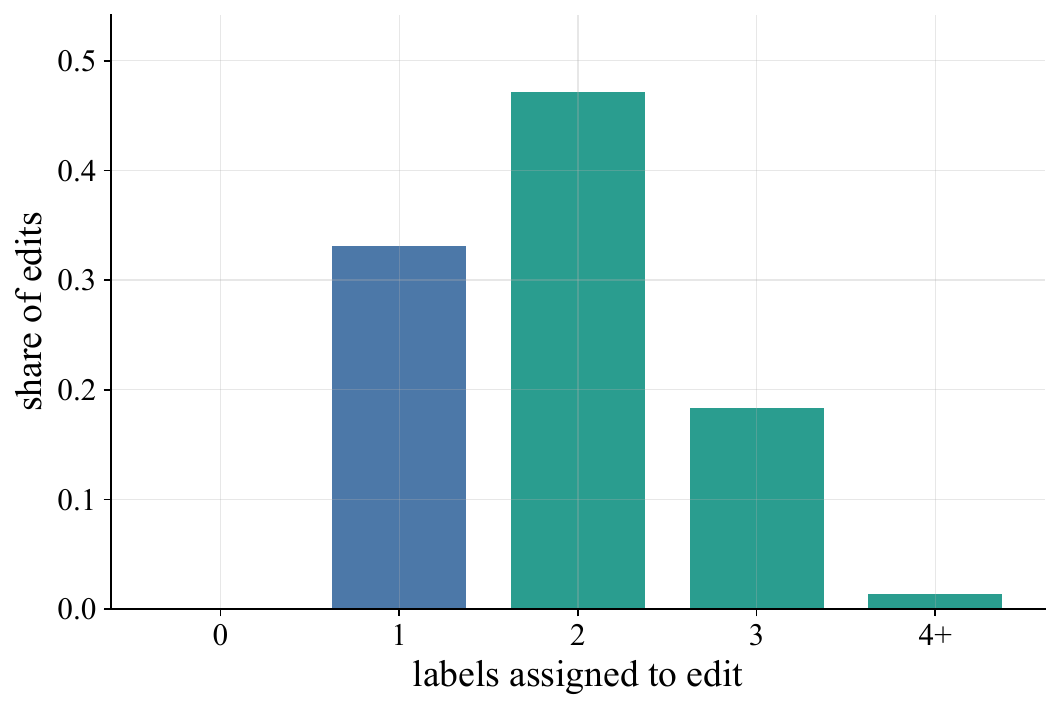}\\
  {\small (b) Math}
\end{minipage}
\caption{\textbf{Distribution of labels per edit, by domain.} Most edits in both domains are multi-label: $52.4\%$ of edits aggregate-wide carry exactly two labels and only $32.4\%$ are single-label. The categories of Figure~\ref{fig:edit-taxonomy-results} should therefore be read as overlapping rather than mutually exclusive modes.}
\label{fig:edit-label-count-by-domain}
\end{figure}

\begin{figure}[h]
\centering
\includegraphics[width=0.85\linewidth]{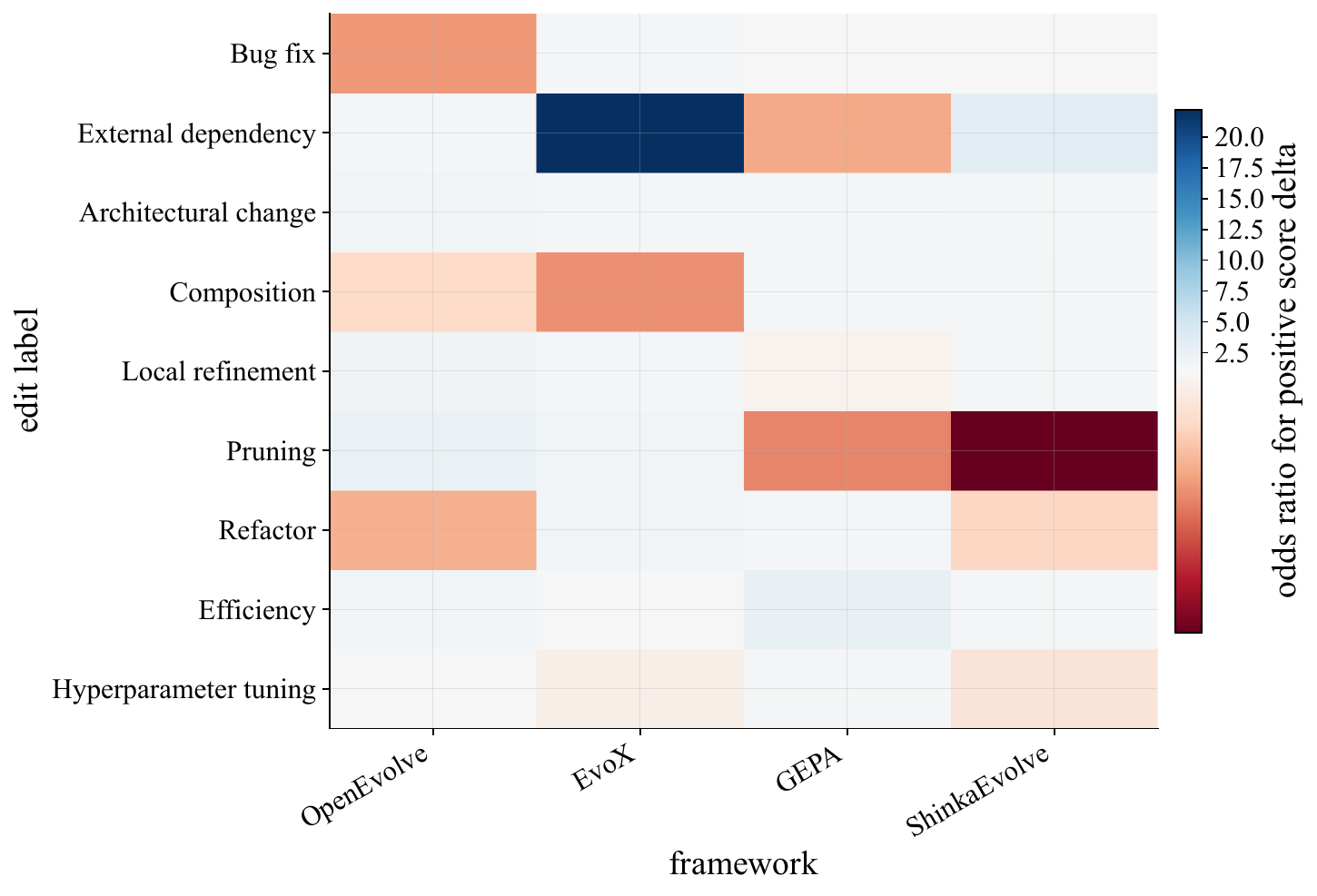}
\caption{\textbf{Per-edit helpfulness by backend.} Odds ratio for positive normalized score change broken down by the four evolutionary backends. Some categories (notably \emph{External dependency}) are consistently positive across backends, while others vary in magnitude. \texttt{openevolve\_native} contributes only $2$ runs to this corpus, so its column should be interpreted with a wider implicit confidence band; we include it for completeness.}
\label{fig:edit-helpfulness-by-backend}
\end{figure}

\subsection{LLM-as-judge validation}
\label{appendix:judge-validation}

\paragraph{Taxonomy origin.}
The 9-category edit taxonomy used in \S\ref{sec:static-analysis} was derived inductively from EvoTrace runs rather than imposed top-down. The first author sampled several classified runs and proposed an initial set of edit categories; these were discussed and refined with co-authors over multiple iterations on further sampled traces, with categories merged or split until the label set stabilised on the nine used to prompt the LLM judge: \emph{hyperparameter\_tuning}, \emph{local\_refinement}, \emph{architectural\_change}, \emph{composition}, \emph{efficiency}, \emph{bug\_fix}, \emph{pruning}, \emph{refactor}, and \emph{external\_dependency}. Table~\ref{tab:edit-type-definitions} gives a one-line working definition for each label; curated example diffs are provided in Appendix~\ref{appendix:edit-examples}.

\begin{table}[h]
\centering
\caption{Working definitions of the nine edit categories. One curated example diff per category is given in Appendix~\ref{appendix:edit-examples}.}
\label{tab:edit-type-definitions}
\small
\renewcommand{\arraystretch}{1.25}
\begin{tabular}{@{}lp{0.72\linewidth}@{}}
\toprule
\textbf{Label} & \textbf{Definition} \\
\midrule
Hyperparameter tuning & Change to one or more numeric literals or configuration values; control flow and surrounding algorithm unchanged. \\
Local refinement      & Targeted edit within an existing routine; the routine's role and structure are preserved. \\
Architectural change  & A core algorithmic block is replaced by a substantively different approach. \\
Composition           & A new component (operator, phase, branch) is added alongside an existing one; existing logic is retained. \\
Efficiency            & Same input--output behaviour reimplemented at lower asymptotic or constant cost (e.g.\ batching, vectorization, partial sorts, caching). \\
Bug fix               & Correction of a latent defect such as a missing guard, wrong sign, off-by-one, or mishandled sentinel. \\
Pruning               & A code path, phase, or feature is removed; the remaining program is kept intact. \\
Refactor              & Behaviour-preserving restructuring: renaming, reordering, extracting helpers, or moving declarations. \\
External dependency   & Introduction or removal of an import or external library. \\
\bottomrule
\end{tabular}
\end{table}

\paragraph{Inter-rater reliability against the LLM judge.}
To validate the LLM-as-judge classifier we conducted a blind multi-label inter-rater reliability study on a stratified sample of $200$ parent$\to$child edits drawn from three classified runs. The first author labelled each edit without seeing the model's output; agreement was then computed against the \texttt{deepseek-chat} judge.

Across the nine categories we observed substantial overall agreement: macro Cohen's $\kappa = 0.77$, mean Jaccard $= 0.86$, micro-$F_1 = 0.90$, and exact-match accuracy $= 74.5\%$.
\subsection{Curated examples per edit type}
\label{appendix:edit-examples}

To make the taxonomy labels of \S\ref{sec:static-analysis} concrete, we illustrate each of the nine categories with one curated example diff drawn from the labeled corpus. Examples are selected for visual clarity rather than for the highest score delta; some carry more than one label (e.g., \emph{Pruning}, \emph{Composition}, \emph{Bug fix}, \emph{Efficiency}, and \emph{External dependency} are each accompanied by other labels in the cleanest paper-ready example), which is itself part of the empirical story analyzed in Appendix~\ref{appendix:edit-taxonomy-breakdowns}: most edits in the corpus are multi-label. Each example below lists run identity, iteration, label set, and score delta; full unified diffs and metadata are bundled with the dataset release.

\paragraph{Hyperparameter tuning.}
\textit{Single cooling-rate change.} \texttt{openevolve\_native} / \texttt{heilbronn\_triangle}, iter $40$, labels $\{\text{hyperparameter\_tuning}\}$, $\Delta s = +0.0153$. The cleanest possible single-knob example: one numeric literal changes; the surrounding algorithm is unchanged.
{\scriptsize
\begin{verbatim}
@@ -74,7 +74,7 @@
     num_restarts = 25            # more restarts to escape local minima
     steps_per_run = 200000       # longer runs for better convergence
     T0 = 0.12                    # higher initial temperature
-    cooling_rate = 0.99992       # slightly slower cooling
+    cooling_rate = 0.99993       # slightly slower cooling (compensate more steps)
     best_min = 0.0
     best_points = None
\end{verbatim}
}

\paragraph{Pruning.}
\textit{Delete final global-shake phase.} \texttt{openevolve\_native} / \texttt{heilbronn\_triangle}, iter $38$, labels $\{\text{hyperparameter\_tuning},\,\text{pruning}\}$, $\Delta s = +0.0670$. A whole final optimization phase is removed (not commented out or renamed); the diff also retunes several literals, so this is a pruning-plus-tuning compound example.
{\scriptsize
\begin{verbatim}
@@ -71,10 +71,10 @@
         return np.min(area)

     # Simulated Annealing parameters
-    num_restarts = 35            # increased restarts to escape local minima
-    steps_per_run = 250000       # longer runs for better convergence
-    T0 = 0.15                    # higher initial temperature for more exploration
-    cooling_rate = 0.99992       # slightly slower cooling
+    num_restarts = 25            # balanced restarts and run length
+    steps_per_run = 280000       # more steps for deeper exploration
+    T0 = 0.12                    # initial temperature
+    cooling_rate = 0.99993       # slightly slower cooling (compensate more steps)
     best_min = 0.0
     best_points = None

@@ -169,24 +169,6 @@
             best_min = current_min_ref
             best_points = points_ref.copy()

-    # Final global shake of best configuration to escape narrow local minima
-    if best_points is not None:
... [truncated, see full diff file] ...
\end{verbatim}
}

\paragraph{Architectural change.}
\textit{Brute-force candidate search replaced by closed-form selection.} \texttt{evox} / \texttt{ale\_bench\_ahc016}, iter $28$, labels $\{\text{architectural\_change},\,\text{local\_refinement}\}$, $\Delta s = +1.37 \times 10^{7}$. The program stops scanning many candidate graph sizes and switches to a derived closed-form rule.
{\scriptsize
\begin{verbatim}
@@ -203,40 +203,49 @@
     double epsilon_noise_rate;
     std::cin >> M_graphs >> epsilon_noise_rate;

-    double best_score = -1e100;
-    int best_N = 4;
-    Strategy best_strat = Strategy::GED;
-
-    // Evaluate GED strategies (N=4,5,6)
-    for (int Ncand : {4,5,6}) {
-        if (Ncand == 6 && M_graphs > 156) continue;
-        if (Ncand == 5 && M_graphs > 34) continue;
-        if (Ncand == 4 && M_graphs > 11) continue;
-        double score = estimate_ged_score(Ncand, M_graphs, epsilon_noise_rate);
-        if (score > best_score) { ... }
-    }
-    // Evaluate edge-count strategies (N=4..100)
-    for (int Ncand = 4; Ncand <= 100; ++Ncand) {
-        double score = estimate_edge_score(Ncand, M_graphs, epsilon_noise_rate);
-        if (score > best_score) { ... }
-    }
+    int N_for_GED_strat;
+    if (M_graphs <= 11) N_for_GED_strat = 4;
... [truncated, see full diff file] ...
\end{verbatim}
}

\paragraph{Local refinement.}
\textit{Add boundary bonus inside the existing scoring heuristic.} \texttt{evox} / \texttt{ale\_bench\_ahc015}, iter $94$, labels $\{\text{local\_refinement}\}$, $\Delta s = +4{,}295$. Small targeted change to a heuristic formula; surrounding algorithm intact.
{\scriptsize
\begin{verbatim}
@@ -283,6 +283,10 @@
                 if ((candy.c == 0 && min_target_c == 0) ||
                     (candy.c == GRID_SIZE - 1 && max_target_c == GRID_SIZE - 1)) {
                      bonus_val += PER_CANDY_BONUS_FACTOR;
+                }
+                // Additional bonus for being at the boundary of the target column range
+                if (candy.c == min_target_c || candy.c == max_target_c) {
+                    bonus_val += 0.5;
                 }
             }
         }
\end{verbatim}
}

\paragraph{Composition.}
\textit{Add swap mutation operator on top of existing plan search.} \texttt{evox} / \texttt{ale\_bench\_ahc026}, iter $57$, labels $\{\text{hyperparameter\_tuning},\,\text{composition}\}$, $\Delta s = +9{,}252$. Main search intact; an additional mutation operator is layered on, alongside several constant retunings.
{\scriptsize
\begin{verbatim}
@@ -18,11 +18,11 @@
 // Constants for heuristic evaluation
-const double HEURISTIC_EMPTY_STACK_BONUS_SCORE = 1500.0;
+const double HEURISTIC_EMPTY_STACK_BONUS_SCORE = 1000.0;
 const double STACK_HEIGHT_PENALTY_FACTOR = 0.1;
 const int    HEURISTIC_LOOKAHEAD_WINDOW = 5;
-const double HEURISTIC_COVER_CRITICAL_PENALTY_PER_BOX_ABOVE = 4.0;
-const double HEURISTIC_MIN_LABEL_IN_DEST_FACTOR = 0.03;
+const double HEURISTIC_COVER_CRITICAL_PENALTY_PER_BOX_ABOVE = 5.0;
+const double HEURISTIC_MIN_LABEL_IN_DEST_FACTOR = 0.05;

@@ -473,7 +467,7 @@
         double op_choice_rand = RGen.a_double(0.0, 1.0);
-        if (op_choice_rand < 0.35 && N_CONST > 0) {
+        if (op_choice_rand < 0.25 && N_CONST > 0) {
... [truncated, see full diff file] ...
\end{verbatim}
}

\paragraph{Bug fix.}
\textit{Check strategy success and INF\_COST sentinel.} \texttt{evox} / \texttt{ale\_bench\_ahc046}, iter $99$, labels $\{\text{bug\_fix},\,\text{hyperparameter\_tuning}\}$, $\Delta s = +56.2$. Parent silently ignored a helper's return value and invalid-cost sentinel; child explicitly guards against both failure modes.
{\scriptsize
\begin{verbatim}
@@ -378,7 +378,7 @@
-const int GREEDY_REOPTIMIZE_SUBSET_SIZE = 40;  // Balanced between exploration and speed
+const int GREEDY_REOPTIMIZE_SUBSET_SIZE = 170; // Full scan for best strategy per segment

@@ -798,12 +798,16 @@
             current_sa_choices[k] = current_best_strategy_code_for_k;

             SegmentExecResult final_segment_res_for_k_build;
-            apply_combined_strategy(current_best_strategy_code_for_k,
+            bool success_seg = apply_combined_strategy(
+                                    current_best_strategy_code_for_k,
                                     player_pos_sim_build,
                                     target_P_k,
                                     greedy_grid_sim_build,
                                     final_segment_res_for_k_build,
                                     true);
+            if (!success_seg || final_segment_res_for_k_build.turns == INF_COST) {
+                possible_greedy = false;
+                break;
+            }
\end{verbatim}
}

\paragraph{Efficiency.}
\textit{\texttt{std::sort} replaced by \texttt{std::nth\_element}.} \texttt{evox} / \texttt{ale\_bench\_ahc027}, iter $46$, labels $\{\text{efficiency},\,\text{hyperparameter\_tuning}\}$, $\Delta s \approx 9.2 \times 10^{18}$. Top-$k$ selection semantics are preserved; full sort is replaced by a partial-ordering primitive.
{\scriptsize
\begin{verbatim}
@@ -266,8 +266,6 @@
              TMP_CELL_DIRT_INFOS_LIST_GLOBAL_BUFFER.push_back(...);
         }
     }
-    std::sort(TMP_CELL_DIRT_INFOS_LIST_GLOBAL_BUFFER.begin(),
-              TMP_CELL_DIRT_INFOS_LIST_GLOBAL_BUFFER.end());

@@ -275,6 +273,13 @@
     // Stochastic: pick uniformly from the top 20% of cells
     int num_candidates = std::max(1,
         (int)TMP_CELL_DIRT_INFOS_LIST_GLOBAL_BUFFER.size() / 5);
+    // Use nth_element to partition: the first num_candidates elements are the largest
+    std::nth_element(TMP_CELL_DIRT_INFOS_LIST_GLOBAL_BUFFER.begin(),
+                     TMP_CELL_DIRT_INFOS_LIST_GLOBAL_BUFFER.begin() + num_candidates,
+                     TMP_CELL_DIRT_INFOS_LIST_GLOBAL_BUFFER.end(),
+                     [](const CellDirtInfo& a, const CellDirtInfo& b) {
+                         return a.weighted_dirt_contribution
+                              > b.weighted_dirt_contribution;
+                     });

@@ -600,7 +605,7 @@
-        const int POST_MAX_ATTEMPTS = 500;
+        const int POST_MAX_ATTEMPTS = 200;
\end{verbatim}
}

\paragraph{External dependency.}
\textit{Introduce JAX and Optax optimization pipeline.} \texttt{shinkaevolve} / \texttt{first\_autocorr\_ineq}, iter $94$, labels $\{\text{architectural\_change},\,\text{external\_dependency}\}$, $\Delta s = +0.991$. The strongest paper example of \emph{external\_dependency}: the imports of \texttt{jax} and \texttt{optax} are unambiguous and visually prominent. Note that the same diff also constitutes a large architectural change, illustrating that this label often appears in combination with others.
{\scriptsize
\begin{verbatim}
@@ -1,3 +1,127 @@
 # EVOLVE-BLOCK-START
-# Paste your original program here. Ensure the evolve block markers are present.
+import jax
+import jax.numpy as jnp
+import optax
+import numpy as np
+from dataclasses import dataclass
+
+@dataclass
+class Hyperparameters:
+    num_intervals: int = 600
+    learning_rate: float = 0.005
+    end_lr_factor: float = 1e-4
+    num_steps: int = 40000
+    warmup_steps: int = 2000
+
+class AutocorrelationOptimizer:
+    def __init__(self, hypers: Hyperparameters):
+        self.hypers = hypers
+        self.domain_width = 0.5
+        self.dx = self.domain_width / self.hypers.num_intervals
... [truncated, see full diff file] ...
\end{verbatim}
}

\paragraph{Refactor.}
\textit{Move \texttt{SimulationResult} definition before use.} \texttt{shinkaevolve} / \texttt{ale\_bench\_ahc026}, iter $68$, labels $\{\text{refactor}\}$, $\Delta s = +9{,}248$. The cleanest refactor diff in the corpus: a type definition is moved to a more appropriate place without changing the algorithm or introducing new logic.
{\scriptsize
\begin{verbatim}
@@ -137,7 +137,13 @@
     }
 };

-// Forward declaration for helper functions
+// Define SimulationResult before any function that uses it
+struct SimulationResult {
+    long long energy_cost;
+    std::vector<std::pair<int, int>> ops_history;
+};
+
+// Forward declaration for helper functions (SimulationResult is now defined)
 std::pair<State, long long> simulate_up_to_k(
     const std::vector<std::vector<int>>& init,
     const std::vector<int>& plan,

@@ -319,11 +325,6 @@
     return run_simulation_from_intermediate_state(
         std::move(st), plan, 0, N, M, record_all);
 }

-struct SimulationResult {
-    long long energy_cost;
-    std::vector<std::pair<int, int>> ops_history;
-};
-
 int main() {
\end{verbatim}
}

\subsection{Cycling: additional analyses}
\label{appendix:cycling-extras}

The body section \S\ref{sec:cycling} reports only the headline cycling result. This appendix covers (i) a finer three-way classifier on the unified diff, (ii) a model- and prompt-dependence breakdown of the tuning share, and (iii) a null result on post-breakthrough cycling that did not survive cross-run aggregation.

\paragraph{Three-way classifier.}
Each parent--child edit is classified deterministically using three categories on the unified diff between $\mathbf{p}_t$ and $p_t$. \emph{Literal recycling}: the added line is byte-identical to a line previously removed elsewhere in the lineage. \emph{Tuning recycling}: the added line's number-collapsed skeleton matches a previously-removed line, but the numeric values differ (coefficient churn). \emph{Trivial recycling}: comment-only or whitespace-only changes. The body cycling rate aggregates all three; per-category rates are emitted by the classifier.

\paragraph{Edit composition is model- and prompt-dependent.}
Restricting to code-changing lines, the median per-run share that the paired-skeleton classifier marks as \emph{tuning recycling} is $8\%$, but the range is wide ($2$--$44\%$). Holding the task fixed at \texttt{ahc015}, the tuning share varies sharply with the generator (Table~\ref{tab:tuning-share-ahc015}). The diff-vs-no-diff axis is the strongest single predictor: the same model (\texttt{deepseek-reasoner}) drops from $20\%$ to $2\%$ tuning share when its diff-based generation is turned off.

\begin{table}[h]
\centering
\caption{Tuning share of code-changing lines on \texttt{ahc015}, holding task fixed. The diff-vs-no-diff axis dominates model identity.}
\label{tab:tuning-share-ahc015}
\small
\setlength{\tabcolsep}{8pt}
\renewcommand{\arraystretch}{1.2}
\begin{tabular}{@{}lc@{}}
\toprule
\textbf{Generator} & \textbf{Tuning share} \\
\midrule
\texttt{gflash}                     & 44\% \\
\texttt{claude-haiku-4-5}           & 26\% \\
\texttt{deepseek-reasoner} (diff)   & 20\% \\
\texttt{deepseek-reasoner} (no diff)& \phantom{0}2\% \\
\bottomrule
\end{tabular}
\end{table}

\paragraph{Negative result on post-breakthrough cycling.}
We initially hypothesized that cycling spikes immediately after a best-so-far event (a refractory period in which the search churns over surrounding code). The change in cycling rate in the $5$ iterations after each best-so-far event has mean $-0.005$, median $-0.021$, and range $[-0.39, +0.23]$ across $26$ runs. Some runs show large positive spikes, others large drops. The hypothesis does not survive cross-run aggregation and we report it as a null.

\subsection{Public-vs-private generalization on ALE}
\label{appendix:public-private}

ALE-bench public scores are not the held-out judging metric. We re-score every ALE run's public best-so-far chain on the private test set used by AtCoder ($n{=}30$ run/problem pairs across the four main backends, covering $10$ ALE problems). Two of the four frameworks overfit on at least $30\%$ of the problems they were scored on, and the same problem can flip generalization sign between frameworks (Table~\ref{tab:public-private-problem}). On \texttt{ahc024}, \texttt{openevolve} found a $+1{,}606$ rating-point private gain (aligned), while \texttt{shinkaevolve}, on the same problem, lost $1{,}610$ rating points despite a positive public score change. On \texttt{ahc027}, three of four frameworks (\texttt{evox}, \texttt{gepa}, \texttt{openevolve}) overfit; only \texttt{shinkaevolve} generalized. Per-framework counts are summarized in Table~\ref{tab:public-private-framework}: \texttt{evox} aligned $0$ of $8$ scored runs, \texttt{gepa} $1$ of $7$, \texttt{openevolve} $4$ of $9$, \texttt{shinkaevolve} $4$ of $6$. The public best-so-far chain is therefore unreliable as a single-number summary of an ALE run; in our data, problem identity is a stronger predictor of overfit than framework identity.

\begin{table}[h]
\centering
\caption{\textbf{Public vs.\ private generalization across frameworks on ALE.} Each cell reports the change in AtCoder rating-point performance from seed to the final public-best program along the run's public best-so-far chain, when re-scored on the held-out private test set. Bold cells flag \emph{overfitting} (public $\uparrow$, private $\downarrow$); ``---'' indicates a single-event lineage, an unscorable seed, or a missing private metric.}
\label{tab:public-private-problem}
\small
\setlength{\tabcolsep}{8pt}
\renewcommand{\arraystretch}{1.15}
% public vs. private ranking on ALE, all 4 frameworks.
% Cell content: \Delta priv. perf. (rating points), seed -> final
% along the public best-so-far chain. Bold = overfit (public up
% but private down). 'aligned' rows show both public and private up.
\begin{tabular}{lrrrr}
\toprule
problem & evox & gepa & openevolve & shinka \\
\midrule
ahc008 & $+0$ & $+0$ & $+37$ & $+38$ \\
ahc011 & \textbf{$-3$} & --- & --- & --- \\
ahc015 & $+0$ & $+0$ & $+0$ & $+2{,}194$ \\
ahc016 & \textbf{$-1$} & $+5$ & $+2$ & $+7$ \\
ahc024 & $+0$ & --- & $+1{,}606$ & \textbf{$-1{,}610$} \\
ahc025 & $+0$ & $+0$ & \textbf{$-16$} & --- \\
ahc026 & $+0$ & $+0$ & $+1{,}735$ & $+10$ \\
ahc027 & \textbf{$-45$} & \textbf{$-489$} & \textbf{$-69$} & $+30$ \\
ahc039 & $+0$ & $+0$ & $+0$ & \textbf{$-32$} \\
ahc046 & $+0$ & $+0$ & \textbf{$-1{,}995$} & $+0$ \\
\bottomrule
\end{tabular}

\end{table}

\begin{table}[h]
\centering
\caption{\textbf{Per-framework counts of aligned vs.\ overfitting public$\to$private trajectories on ALE.} A run is \emph{aligned} if both public and private scores improved from seed to final, \emph{mild overfit} if private worsened by $\le 200$ rating points despite a public gain, and \emph{severe overfit} if by more than $200$ points; the remainder is no movement on private.}
\label{tab:public-private-framework}
\small
\setlength{\tabcolsep}{8pt}
\renewcommand{\arraystretch}{1.15}
% Per-framework summary: how often does the public BSF chain
% overfit (public up but private down) on ALE?
\begin{tabular}{lrrrr}
\toprule
framework & runs scored & aligned & overfit (mild) & overfit (severe) \\
\midrule
evox & 8 & 0 & 3 & 0 \\
gepa & 7 & 1 & 0 & 1 \\
openevolve & 9 & 4 & 2 & 1 \\
shinka & 6 & 4 & 1 & 1 \\
\bottomrule
\end{tabular}

\end{table}

\subsection{A walk-through of one short-span cycle}
\label{appendix:cycling-example}

To make the deterministic cycling classifier of \S\ref{sec:cycling} concrete, we trace one short-span cycle in a single \texttt{openevolve\_native} / \texttt{heilbronn\_triangle} run. At iteration $i$ the parent contains a hand-tuned annealing schedule with \texttt{cooling\_rate = 0.99992}; the child at $i{+}1$ rewrites the schedule, \emph{deletes} the line, and replaces it with \texttt{cooling\_rate = 0.99988}. By iteration $i{+}5$ the search has produced a child whose diff against its own parent re-adds the byte-identical line \texttt{cooling\_rate = 0.99992}, classified as \textbf{literal recycling} by the paired-skeleton classifier (the added line matches a previously-removed line in the lineage exactly, including the trailing comment). The same constant is deleted again at $i{+}9$ and re-introduced at $i{+}14$, and so on; the cycle is short-span (median across all classified runs: 5 iterations between deletion and re-introduction) and accumulates over the run, contributing to the monotonic per-iteration cycling-rate growth reported in \S\ref{sec:cycling}.

\subsection{Bayesian optimization baseline}
\label{appendix:bo-details}

This section documents the BO baseline used for the tuning-gap analysis in \S\ref{sec:tuning-gap}. The goal is to estimate $f^{\star}_{\mathrm{BO}}(s_0)$ for a fixed seed structure $s_0$ by tuning only its embedded numeric constants, with no further structural search. The pipeline has three stages: (i) a single LLM call that identifies tunable knobs and proposes intervals, (ii) a deterministic rewrite that exposes those knobs as a top-level parameter block, and (iii) a short \texttt{gp\_minimize} run that calls the original evaluator harness.

\paragraph{Knob identification (one LLM call).}
We send the program source to \texttt{deepseek-reasoner} via an OpenAI-compatible chat endpoint and ask for a JSON list of candidate knobs. There is no agentic loop and no retry: a single structured-output call returns a list of objects with fields \texttt{name}, \texttt{source\_literal}, \texttt{context\_line}, \texttt{default}, \texttt{low}, \texttt{high}, \texttt{scale} (\texttt{linear} or \texttt{log}), \texttt{kind} (\texttt{int} or \texttt{float}), and a one-sentence \texttt{rationale}. The system prompt restricts candidates to solver tolerances, iteration/sample budgets, step sizes and learning rates, soft penalty/reward weights, cooling rates, and threshold dispatches; problem constants (population sizes that the evaluator reads, grid dimensions, the $n$ in $n$-circle packing), mathematical identities ($\sqrt{2}$, $\pi$), array shapes, and boolean flags are explicitly excluded. Range guidelines are calibrated by parameter family (log-scale spans of $10^{2}$ for step sizes and weights, $10^{3}$ for tolerances; linear $[0, 1]$ for probabilities and acceptance ratios; etc.), with a hard rule that any range with $\mathrm{high}/\mathrm{low} \ge 100$ uses a log scale. The prompt caps the proposal at 8 knobs, since BO budget scales poorly past that, and instructs the model to skip any literal that appears multiple times in the source (ambiguous replacement target).

\paragraph{Validation and rewrite.}
Each returned spec is validated: the \texttt{source\_literal} must appear byte-identically in the source, and the \texttt{context\_line} must match a line in the file. For Python targets, a \texttt{PARAMS = \{...\}} dict is injected at the top of the file (after shebang/encoding/docstring) and each accepted literal is replaced in its context line by \texttt{PARAMS["name"]}. For C++ targets, a block of \texttt{\#define \_BO\_NAME value} macros is inserted after the last \texttt{\#include} and the literal is replaced by the corresponding macro. Replacement uses a regex that excludes longer-number neighbors (so \texttt{0.1} does not match inside \texttt{0.123}). Specs whose literal can be found in the source but not in the proposed context line are dropped without retry; the BO then runs over the surviving knobs only, so the effective per-target knob count (median $6$ in the runs reported in \S\ref{sec:tuning-gap}) is generally smaller than the LLM's nominal proposal.

\paragraph{BO loop.}
Tuning runs use \texttt{skopt.gp\_minimize} (scikit-optimize) with 24 evaluator calls per target: 8 random initial points followed by 16 BO acquisitions over a Gaussian-process surrogate. Each call substitutes the current parameter vector into the rewritten source (regenerating the \texttt{PARAMS} dict or rewriting the \texttt{\#define} block) and invokes the same evaluator harness used during the original evolutionary run, so $f^{\star}_{\mathrm{BO}}(s_0)$ and $f^{\star}_{\mathrm{evo}}$ are directly comparable. Reported numbers in \S\ref{sec:tuning-gap} (Table~\ref{tab:bo-outcome}) are best-of-24 minus the seed program's score $f(p_0)$.

\paragraph{Knob-identification prompt.}
The system prompt sent to \texttt{deepseek-reasoner} is reproduced verbatim below; the user message is just the program source wrapped in a fenced block tagged with the language.

\small
\begin{verbatim}
You are an expert at identifying tunable hyperparameters in heuristic
optimisation code.

Your task: read a program and return a JSON list of numeric constants
that are good candidates for Bayesian-optimisation tuning, WITH
SENSIBLE RANGES.

GOOD candidates:
  - Solver tolerances / convergence thresholds (e.g. 1e-6, 0.001)
  - Iteration / restart / sample budgets (e.g. 1000, 50, 10)
  - Step sizes, perturbation scales, learning rates
  - Soft penalty / reward weights (e.g. 0.5, 2.0)
  - Cooling / annealing rates (e.g. 0.95, 0.99)
  - Threshold dispatches (e.g. `if depth < 30: return X else return Y`)

BAD candidates (do NOT include):
  - Problem constants (n=26, n_circles=26, dim=2, GRID_SIZE=10)
  - Mathematical identities (sqrt(2), pi, e)
  - Array shapes / index bounds
  - Evaluator-facing constants (timeout values that the harness reads)
  - Constants inside identities the model should preserve
    (vertices, basis vectors)
  - True boolean flags (use=True)

For EACH selected knob, return:
  - name           : a unique snake_case identifier (e.g. "step_size")
  - source_literal : the EXACT numeric literal as it appears in the
                     source (e.g. "0.1", "1e-6", "0.95"). Must be
                     byte-identical.
  - context_line   : the line of code where the literal appears
                     (verbatim, used as a disambiguator for replacement)
  - default        : the default value (= source_literal as a number)
  - low / high     : interval bounds for BO (see RANGE GUIDELINES)
  - scale          : "linear" or "log" (see RANGE GUIDELINES)
  - kind           : "int" or "float"
  - rationale      : one short sentence

RANGE GUIDELINES -- cover orders of magnitude when the parameter family
warrants it. Don't propose timid +-20% intervals: BO can only discover
what the search space contains. Use these defaults:

  parameter family            scale   range relative to default
  --------------------------  ------  ---------------------------------
  learning rate / step size   log     [default/100, default*100]
  temperature / amplitude     log     [default/100, default*100]
  tolerance / threshold       log     [default/1000, default*1000]
  penalty / reward weight     log     [default/100, default*100]
  cooling factor (0<x<1)      linear  [0.5, 0.99999]   (always wide)
  acceptance prob / sigmoid   linear  [0.0, 1.0]
  restart count / pop size    log     [max(1, default/10), default*10]
  iteration budget            log     [default/5, default*5]
  sample count                log     [max(1, default/10), default*10]
  small-int categorical       linear  [1, max(default*5, 20)]

If a parameter doesn't fit a family above, default to:
  - log scale if default is a positive non-integer < 1 or > 100
  - linear scale otherwise
  - range that spans at least one order of magnitude (high/low >= 10)

HARD RULES:
  - If high/low >= 100, scale MUST be "log" (BO converges much faster
    on log-scale priors over wide intervals).
  - Cooling factors and probabilities stay within their natural [0, 1]
    range even if that constrains low/high.
  - For integers: low = max(1, floor(low_proposed));
                  high = ceil(high_proposed).

CONSERVATIVE RULES:
  - Pick at most 8 knobs. Fewer is fine -- quality over quantity.
  - If a literal appears multiple times in the source, skip it
    (ambiguous).
  - If unsure whether a constant is tunable, skip it.

Return ONLY a JSON object of the form:
    {"hparams": [ { ...one knob... }, ... ]}
No commentary, no markdown fences.
\end{verbatim}
\normalsize

\end{document}